\documentclass{article} 
\usepackage[T1]{fontenc}
\usepackage{iclr2025_conference,times}

\usepackage{amsmath,amsfonts,bm}

\def\eqref#1{equation~\ref{#1}}

\def\1{\bm{1}}

\DeclareMathAlphabet{\mathsfit}{\encodingdefault}{\sfdefault}{m}{sl}
\SetMathAlphabet{\mathsfit}{bold}{\encodingdefault}{\sfdefault}{bx}{n}

\usepackage[breaklinks=true]{hyperref}
\usepackage{breakurl}
\usepackage{graphicx}
\usepackage{wrapfig}
\usepackage{placeins}
\usepackage{booktabs}
\usepackage{multirow}
\usepackage{enumitem}
\usepackage{float}
\usepackage{tabularx}
\PassOptionsToPackage{table,usenames,dvipsnames}{xcolor}
\usepackage{xcolor}
\usepackage{graphicx}
\usepackage[most]{tcolorbox}
\usepackage{lipsum} 
\usepackage{anyfontsize} 
\definecolor{glaucous}{rgb}{0.38, 0.51, 0.71}
\definecolor{pastelmagenta}{rgb}{0.96, 0.6, 0.76}
\definecolor{orchid}{rgb}{0.85, 0.44, 0.84}

\hypersetup{
  colorlinks=true,
  citecolor=glaucous,
  linkcolor=pastelmagenta,
  urlcolor=orchid}

\usepackage{arydshln}
\makeatletter
\def\adl@drawiv#1#2#3{%
        \hskip.5\tabcolsep
        \xleaders#3{#2.5\@tempdimb #1{1}#2.5\@tempdimb}%
                #2\z@ plus1fil minus1fil\relax
        \hskip.5\tabcolsep}
\newcommand{\cdashlinelr}[1]{%
  \noalign{\vskip 1.3pt
           \global\let\@dashdrawstore\adl@draw
           \global\let\adl@draw\adl@drawiv}
  \cdashline{#1}[.4pt/2pt]
  \noalign{\global\let\adl@draw\@dashdrawstore
           \vskip 3pt}}
\makeatother

\usepackage{marvosym}

\title{\centering EuroLLM-9B: Technical Report}

\author{
\vspace{0.3cm}
\bf
 Pedro Henrique Martins*\textsuperscript{\Neptune} \hspace{0.1cm}
 João Alves*$^{1}$ \hspace{0.1cm}
 Patrick Fernandes*$^{2,3}$ \hspace{0.1cm}
 Nuno M. Guerreiro*$^{1,2,4}$ 
 \\
\bf
 Ricardo Rei*$^{1}$ \hspace{0.1cm}
 Amin Farajian*$^{1}$ \hspace{0.1cm}
 Mateusz Klimaszewski*$^{6}$ \hspace{0.1cm}
 Duarte M. Alves$^{2}$ \hspace{0.1cm}
 \\
\bf
 José Pombal$^{1,2}$ \hspace{0.1cm}
 Nicolas Boizard$^{4,10}$ \hspace{0.1cm}
 Manuel Faysse$^{4,5}$ 
 Pierre Colombo$^{4,7}$ \hspace{0.1cm}
 François Yvon${^9}$ \hspace{0.1cm}
 \\
\bf
 Barry Haddow$^{6,8}$ \hspace{0.1cm}
José G. C.~de Souza$^{1}$ \hspace{0.1cm}
 Alexandra Birch$\diamond^{6,8}$ \hspace{0.1cm}
 André F. T. Martins$\diamond^{1,2}$
\\
\vspace{0.3cm}
\normalfont
$^{1}$Unbabel \hspace{0.1cm} $^{2}$Instituto de Telecomunicações \& Instituto Superior Técnico, Universidade de Lisboa \\ 
$^{3}$Carnegie Mellon University \hspace{0.1cm} $^{4}$MICS, CentraleSupélec, Université Paris-Saclay\\
$^{5}$Illuin Technology \hspace{0.1cm} $^{6}$University of Edinburgh \hspace{0.1cm}  $^{7}$Equall \hspace{0.1cm} $^{8}$Aveni \\
$^{9}$Sorbonne Université, CNRS, ISIR \hspace{0.1cm} $^{10}$Diabolocom
}

\newcommand\blfootnote[1]{%
  \begingroup
  \renewcommand\thefootnote{}\footnote{#1}%
  \addtocounter{footnote}{-1}%
  \endgroup
}

\iclrfinalcopy 

\begin{document}

\maketitle

\begin{abstract}
\blfootnote{* Core contributors, $\diamond$ equal contributors, \textsuperscript \Neptune work done while working at Unbabel}

This report presents \emph{EuroLLM-9B}, a large language model trained from scratch to support the needs of European citizens by covering all 24 official European Union languages and 11 additional languages. EuroLLM addresses the issue of European languages being underrepresented and underserved in existing open large language models. We provide a comprehensive overview of EuroLLM-9B's development, including tokenizer design, architectural specifications, data filtering, and training procedures. We describe the pre-training data collection and filtering pipeline, including the creation of \textit{EuroFilter}, an AI-based multilingual filter, as well as the design of \textit{EuroBlocks-Synthetic}, a novel synthetic dataset for post-training that enhances language coverage for European languages.

Evaluation results demonstrate EuroLLM-9B's competitive performance on multilingual benchmarks and machine translation tasks, establishing it as the leading open European-made LLM of its size. To support open research and adoption, we release all major components of this work, including the base and instruction-tuned models, the EuroFilter classifier, and the synthetic post-training dataset.\footnote{Resources available in HuggingFace as part of the  \href{https://huggingface.co/collections/utter-project/eurollm-66b2bd5402f755e41c5d9c6d}{EuroLLM Collection}.}
\end{abstract}

\section{Introduction}
Large language models (LLMs) have emerged as key drivers of progress in natural language processing (NLP) and artificial intelligence (AI), with notable examples including OpenAI’s GPT series~\citep{openai2024gpt4technicalreport}, Anthropic's Claude~\citep{ClaudeThree} or Google's Gemini~\citep{geminiteam2025geminifamilyhighlycapable}. LLMs are first pre-trained on vast amounts of unlabelled data relying on a self-supervised task (\textit{e.g.,} next word prediction or missing word prediction). This process enables the model to acquire knowledge, to develop strong language understanding and generation skills, and to perform various downstream tasks, often leveraging in-context learning techniques. 
Following pre-training, LLMs are further refined through post-training techniques that enhance their ability to follow natural language instructions, improve task-specific performance, and enhance the adherence to safety protocols.

Despite the growing availability of open-weight LLMs such as LLaMA, Mistral, Gemma, DeepSeek, and Qwen \citep{dubey2024llama, jiang2023mistral, team2024gemma, deepseekai2025deepseekv3technicalreport, qwen2025qwen25technicalreport}, most advanced models are closed, owned by major corporations with limited commitment to open science.
Moreover, existing open models primarily focus on English and a few high-resource languages, leaving many European languages underserved.

To address this gap, we launched the \href{https://eurollm.io/}{EuroLLM project} with the aim of developing a suite of open LLMs capable of understanding and generating text in all 24 official European Union languages (Bulgarian, Croatian, Czech, Danish, Dutch, English, Estonian, Finnish, French, German, Greek, Hungarian, Irish, Italian, Latvian, Lithuanian, Maltese, Polish, Portuguese, Romanian, Slovak, Slovenian, Spanish, and Swedish). 
As the aim of EuroLLM is to provide EU citizens with powerful and useful AI tools, it is critical that the model can also translate and answer questions in other European and non-European languages. With this in mind, we added support for 11 additional languages (Arabic, Catalan, Chinese, Galician, Hindi, Japanese, Korean, Norwegian, Russian, Turkish, and Ukrainian). 
Our journey began with the release of EuroLLM-1.7B \citep{martins2024eurollm}, a compact yet powerful model that excels in machine translation and performs competitively on general benchmarks.
Building on this foundation, we now introduce the technical report on EuroLLM-9B, a model that, at the date of its release (December 9, 2024), was the most capable open European-made LLM of its size.

This technical report provides a comprehensive overview of the development and evaluation of EuroLLM-9B:
\begin{itemize}
    \item We begin by examining the pre-training phase of EuroLLM in \S\ref{sec:pre_training}, where we introduce and describe \textit{EuroFilter}---a multilingual AI-based filter used to curate the pre-training data. We also release EuroFilter alongside this report to support future work on high-quality multilingual data filtering.
    \vspace{0.1cm}
    \item In \S\ref{sec:post_training}, we describe the post-training process of EuroLLM-9B and introduce \textit{EuroBlocks-Synthetic}---a post-training dataset that extends existing synthethic data with multilingual instructions covering many European languages.
    \vspace{0.1cm}
    \item Finally, we present EuroLLM-9B's performance on multilingual benchmarks in \S\ref{sec:results}, comparing it with leading open LLMs.
\end{itemize} 

To support further research and development, we openly release all major components introduced in this report: the base model (\href{https://huggingface.co/utter-project/EuroLLM-9B}{EuroLLM-9B}), the instruction-tuned variant (\href{https://huggingface.co/utter-project/EuroLLM-9B-Instruct}{EuroLLM-9B-Instruct}), the multilingual data filter (\href{https://huggingface.co/utter-project/EuroFilter-v1}{EuroFilter}), and the synthetic post-training dataset (\href{https://huggingface.co/datasets/utter-project/EuroBlocks-SFT-Synthetic-1124}{EuroBlocks-Synthetic}).

\section{Pre-training}

\label{sec:pre_training}
In this section, we describe the pre-training process of EuroLLM-9B, covering the tokenizer modeling (\S\ref{sec:tokenizer}), architectural decisions (\S\ref{sec:pre_training_modeling}), the pre-training phases (\S\ref{sec:pre_training_phases}), and the pre-training data (\S\ref{sec:pre_training_data}). We carried out the pre-training using the Megatron-LM codebase~\citep{shoeybi2020megatronlmtrainingmultibillionparameter}.
\subsection{Tokenizer}
\label{sec:tokenizer}

The starting point for developing EuroLLM is designing the tokenizer that best fits our language coverage. To train the tokenizer, we adopt the BPE~\citep{sennrich-etal-2016-neural} with byte-fallback algorithm, following the approach used by the LLaMa-2 and Mistral-7B models~\citep{touvron2023llama,jiang2023mistral}. To do so, we use the SentencePiece framework~\citep{kudo2018sentencepiece}. 
For multilingual language models, the  vocabulary size of the tokenizer presents a crucial trade-off: while larger vocabularies enable efficient processing across multiple languages, they also increase the model's embedding parameter count. Through experimentation, we determined that a vocabulary size of 128,000 pieces offers a good balance between these competing factors. 

\begin{figure}[h]
    \centering
    \includegraphics[width=\textwidth]{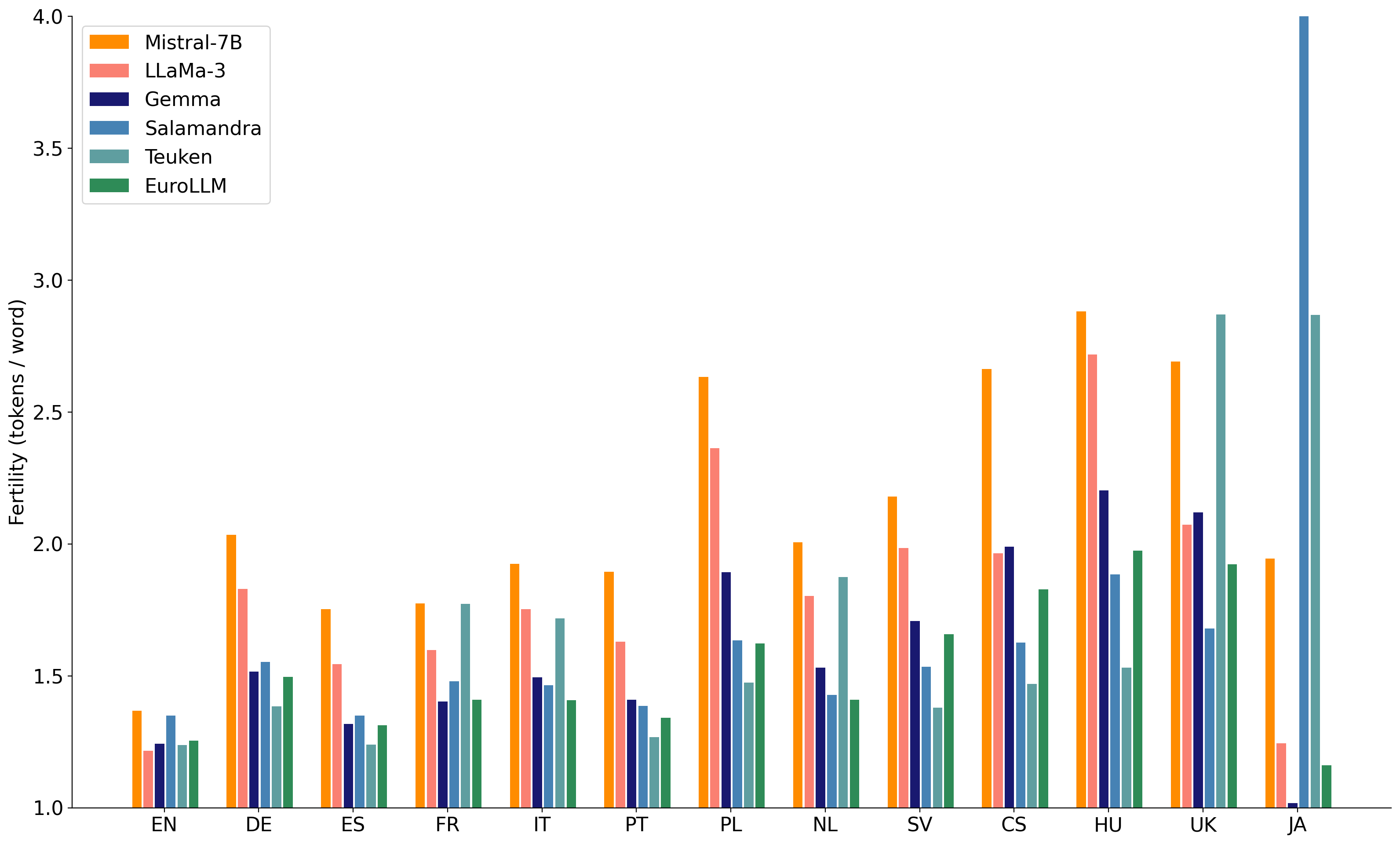}
    \caption{Fertility (tokens per word) obtained with the Mistral-7B, LLaMa-3, Gemma-2, Salamandra, Teuken, and EuroLLM tokenizers for a subset of the EuroLLM languages. With 128k tokens, EuroLLM achieves comparable fertilities to 256k token models (Gemma, Teuken, Salamandra) for most languages, saving 50\% of the embedding parameters.}
    \label{fig:fertility_comparison}
\end{figure}

To evaluate our tokenizer's performance, we conducted a comparison with the tokenizers of several open-weight LLMs: Mistral-7B, LLaMa-3, Gemma-2, Teuken, and Salamandra \citep{jiang2023mistral,llama3modelcard,team2024gemma,ali2024teuken,gonzalezagirre2025salamandratechnicalreport}.
The comparison focuses on tokenizer fertility---the average number of tokens per word. The compared models feature varying vocabulary sizes: Mistral-7B with 32,000 pieces, LLaMa-3 with 128,256 pieces, Teuken with 250,680 pieces, and both Gemma-2 and Salamandra with 256,000 pieces. For this analysis, we used a concatenation of the \textsc{Flores-200} \citep{nllb2022} and Universal Dependencies datasets \citep{nivre2020universal} for each language.

Figure \ref{fig:fertility_comparison} presents the fertility rates for a subset of the languages included in EuroLLM.  Our analysis reveals that, compared to the Mistral-7B tokenizer, the larger vocabulary of EuroLLM leads to significantly lower fertility rates. While LLaMa-3, with its similar vocabulary size, demonstrates superior (lower) fertility for English, it shows higher fertility rates for most other languages in our evaluation.
Notably, despite having a smaller vocabulary than Gemma-2, Teuken, and Salamandra, EuroLLM achieves comparable fertility levels across the evaluated languages.

\subsection{Modeling}
\label{sec:pre_training_modeling}

\begin{wraptable}[19]{r}{0.5\textwidth}
\small
\vspace{-0.3cm}
\setlength{\tabcolsep}{1.5ex}
\begin{tabular}{lc}
\toprule
         & \textbf{9B} \\ \midrule
Sequence Length & 4,096 \\
Number of Layers & 42 \\
Embedding Size & 4,096 \\
FFN Hidden Size & 12,288 \\
Number of Heads & 32 \\
Number of KV Heads (GQA) & 8 \\
Activation Function & SwiGLU \\
Position Encodings & RoPE ($\Theta$=10,000) \\
Layer Norm & RMSNorm \\
Tied Embeddings & No \\
Max Learning Rate & $3\times 10^{-4}$ \\
\midrule
Embedding Parameters & 0.524B \\
LM Head Parameters & 0.524B \\
Non-embedding Parameters & 8.105B \\
Total Parameters & 9.153B \\
\bottomrule
\end{tabular} 
\caption{EuroLLM-9B hyperparameters.}
\label{tab:hyperparameters}
\end{wraptable}
EuroLLM-9B uses a standard, dense Transformer architecture \citep{vaswani2017attention} with the same design choices as EuroLLM-1.7B~\citep{martins2024eurollm}.
\begin{itemize}[leftmargin=0.5cm, itemsep=0pt]
    \item We use grouped query attention (GQA; \cite{ainslie2023gqa}) with 8 key-value heads, which has been demonstrated to enhance inference speed while preserving downstream performance \citep{team2024gemma2}.
    \vspace{0.15cm}
    \item For improved training stability, we use pre-layer normalization  \citep{xiong2020layer} and RMSNorm \citep{zhang2019root}, which offers faster computation compared to LayerNorm \citep{ba2016layer}.
    \vspace{0.15cm}
    \item We use the SwiGLU activation function \citep{shazeer2020glu} since it has been shown to lead to good results on downstream tasks \citep{shazeer2020glu,le2022language}.
    \vspace{0.15cm}
    \item We use rotary positional embeddings (RoPE; \cite{su2024roformer}) in every layer since these have been shown to lead to good performances while allowing the extension of the context length \citep{chen2023extending}.
\end{itemize}

\subsection{Pre-training Phases}
\label{sec:pre_training_phases}
\begin{figure}[h]
    \centering
    \includegraphics[width=14.5cm]{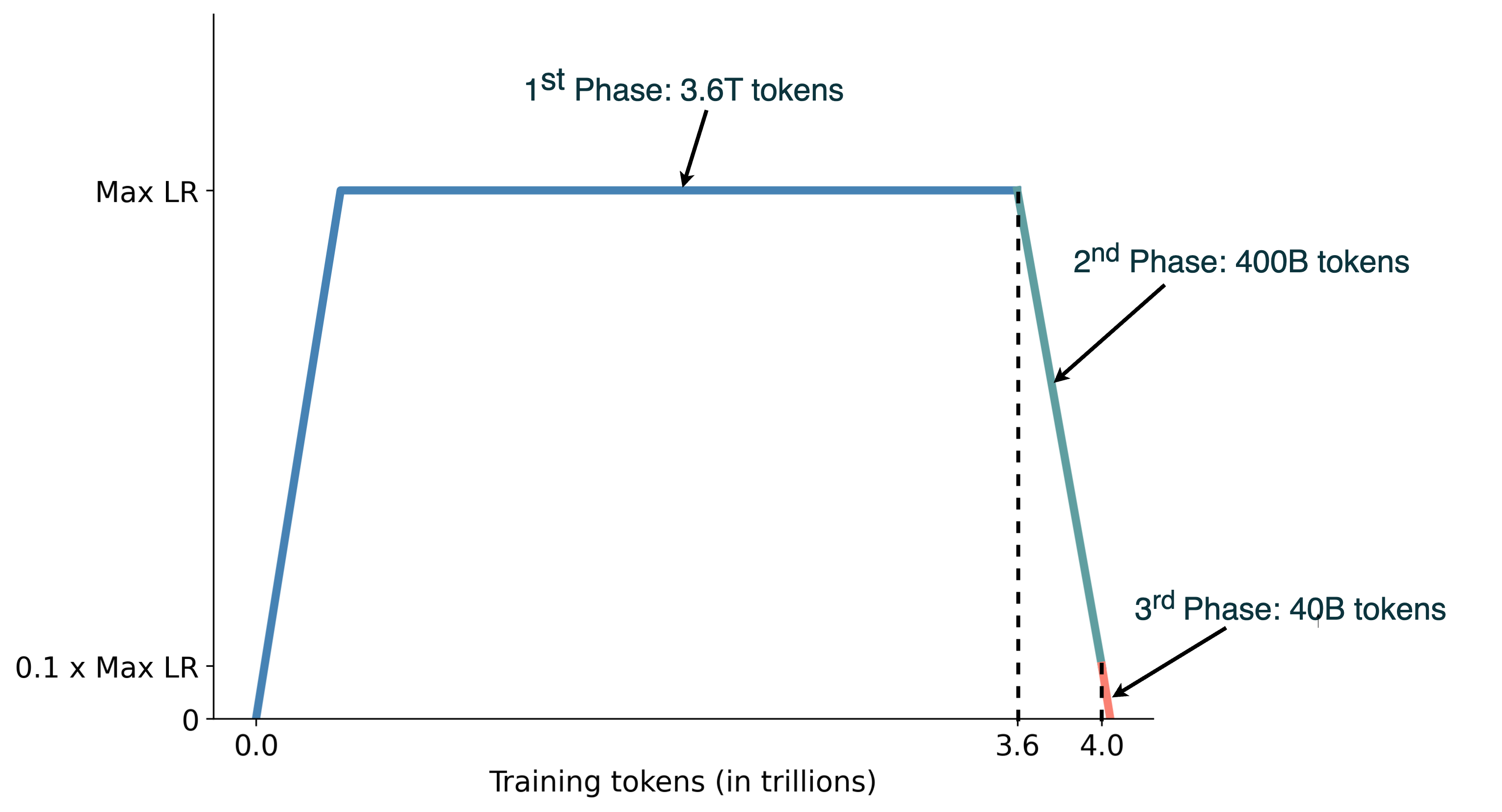}
    \caption{Scheme of the learning rate scheduler.}
    \label{fig:lr_scheduler}
\end{figure}
We pre-train EuroLLM-9B on approximately 4 trillion tokens using a trapezoid learning rate scheduler \citep{xing2018walk} (also named Warmup-Stable-Decay \citep{hu2024minicpm}). We conduct training on 400 Nvidia H100 GPUs from the MareNostrum 5 supercomputer, maintaining a constant batch size of 2,800 sequences (approximately 12 million tokens), and employing the Adam optimizer \citep{kingma2014adam}, with \texttt{bfloat16} mixed precision.

The training process consists of three distinct phases, as shown in Figure~\ref{fig:lr_scheduler}. As we progress through training, we expose the model to higher quality and specialised data~\citep{llama3modelcard}.
\begin{itemize}
    \item \textbf{1\textsuperscript{st} Phase: Learning Rate Warm-up and Plateau}. The initial phase involves linearly increasing the learning rate during the first 10\% of training steps followed by maintaining it at a constant level for the subsequent 80\%. This phase comprises approximately 3.6 trillion tokens.
    \vspace{0.15cm}
    \item \textbf{2\textsuperscript{nd} Phase: Learning Rate Annealing}. During this phase, the learning rate decays linearly from its maximum value ($3\times 10^{-4}$) to 10\% of its peak ($3\times 10^{-5}$). This annealing period encompasses roughly 10\% of training steps, processing about 400 billion tokens.
    \vspace{0.15cm}
    \item \textbf{3\textsuperscript{rd} Phase: Annealing to Zero}. During the final phase, the learning rate decays linearly to zero over a brief period, corresponding to about 40 billion tokens. This phase is a novelty of EuroLLM-9B compared to EuroLLM-1.7B~\citep{martins2024eurollm}.
\end{itemize}

We will describe next the data used in these three phases.

\subsection{Data}
\label{sec:pre_training_data}

\begin{figure}[h]
    \centering
    \includegraphics[width=\linewidth]{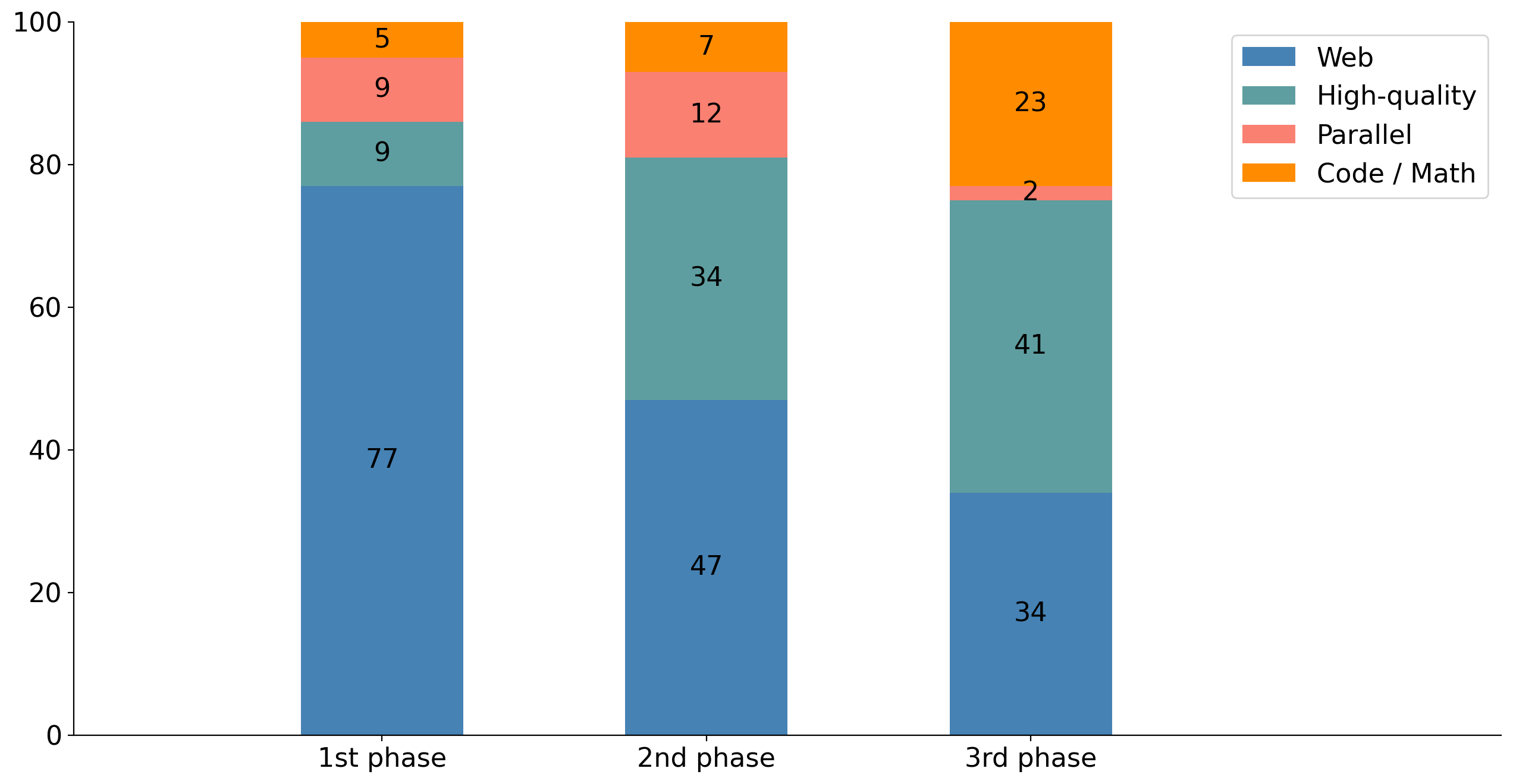}
    \caption{Percentage attributed to each data category in the 3 pre-training phases.}
    \label{fig:percentage_category}
\end{figure}

To train EuroLLM-9B, we collect and filter data from various sources for all supported languages. The data that composes the final corpus can be categorized into four main types: web data, parallel data, code / math data, and higher-quality data. The data distribution for the first two phases is kept the same as that of the 1.7B model. Figure~\ref{fig:percentage_category} illustrates the distribution of these categories across the three pre-training phases.

\subsubsection{Data Collection and Filtering~(EuroFilter)}
\paragraph{\textbf{Web Data}}
For web data collection, we employ different strategies based on language resources.
For English, we use the FineWeb-edu dataset \citep{lozhkov2024fineweb-edu}, selecting documents with educational scores above 2 according to their model-based classifier. The dataset underwent individual dump deduplication and heuristic filtering.

For other high-resource languages~(German, Spanish, French, and Italian), we collect data from RedPajama-Data-v2 \citep{together2023redpajama}, which has been pre-deduplicated.  We further apply perplexity filtering using KenLM \citep{heafield-2011-kenlm}, along with various heuristic filters. Specifically, we remove documents with fewer than 200 characters \citep{mt5-xue2021mt5massivelymultilingualpretrained} and any page containing the phrase “lorem ipsum,” the word “javascript,” or curly brackets \citep{t5-raffel2023exploringlimitstransferlearning}. Additionally, we exclude paragraphs where the fraction of uppercase letters exceeds 40\%, the symbol-to-word ratio is greater than 0.1, or the ratio of words without alphabetic letters exceeds 0.2 \citep{gopher-rae2022scalinglanguagemodelsmethods}.

For the remaining languages, we aggregate data from several datasets: HPLT \citep{degibert2024}, MADLAD-400 \citep{kudugunta2023}, CulturaX \citep{nguyen2023}, and mC4 \citep{xue2021}. After concatenation, we perform deduplication, language identification, perplexity filtering, and the same set of heuristic filters that we used for the high-resource languages, using a CCNet-based preprocessing pipeline \citep{wenzek2019}.

To further improve the quality of the web data used in the 2\textsuperscript{nd} and 3\textsuperscript{rd} training phases, we raise the FineWeb-edu score threshold to 3 for English. For the other languages, we reuse the FineWeb-Edu~\citep{lozhkov2024fineweb-edu} annotation; however, we translate all data using \textsc{Tower v2}-supported languages~\citep{rei-etal-2024-tower}. This is done to create multilingual texts paired with educational scores. Then we train a multilingual classifier on top of mDeBERTa~\citep{he2023debertav} which we use to annotate the rest of the languages. Our filter is publicly available at \href{https://huggingface.co/utter-project/EuroFilter-v1}{utter-project/EuroFilter-v1}.





\paragraph{\textbf{Parallel Data}}
Regarding parallel data, we collect sentence-level to-English (xx→en) and from-English (en→xx) parallel data from various public sources listed in Table~\ref{tab:parallel_data_sources}.

We use Bifixer \citep{prompsit:2020:EAMT} to remove duplicates and ensure translation quality by removing sentence pairs below quality thresholds for Bicleaner \citep{prompsit:2018:WMT,prompsit:2020:EAMT} and \textsc{CometKiwi-22} \citep{rei-etal-2022-cometkiwi}. For Bicleaner, we use a threshold of 0.6 for Portuguese and of 0.5 for all the other languages. For \textsc{CometKiwi-22} we use a threshold of 0.7.

For the 2\textsuperscript{nd} and 3\textsuperscript{rd} training phases we also collect document-level data from Europarl \citep{koehn-2005-europarl} and ParaDocs \citep{paradocs}, applying the same filtering criteria.

\begin{table}[h]
\begin{center} 
\footnotesize
\setlength{\tabcolsep}{5ex}
\begin{tabular}{@{\hspace{0.2cm}}lccc}
\toprule
Dataset         & Version   &  \\ \midrule
Europarl \citep{koehn-2005-europarl}      & v8                 &  \\
ParaCrawl \citep{espla-etal-2019-paracrawl}       & v9                  &  \\
MultiParaCrawl \citep{espla-etal-2019-paracrawl}  & v7.1               &  \\
CCMatrix \citep{schwenk2020ccmatrix}       & v1                &  \\
CCAligned \citep{el-kishky-etal-2020-ccaligned}      & v1               &  \\
MultiCCAligned \citep{el-kishky-etal-2020-ccaligned} & v1              &  \\
WikiTitles \citep{tiedemann2012opus}     & v2014           &  \\
WikiMatrix \citep{schwenk2019wikimatrix}     & v1              &  \\
News-Commentary \citep{tiedemann2012opus} & v16                 &  \\
OPUS100 \citep{zhang2020improving} & v1                 &  \\
TildeModel \citep{rozis-skadins-2017-tilde}     & v2018          &  \\
Bible \citep{mayer-cysouw-2014-creating}          & v1               &  \\
Ubuntu \citep{tiedemann2012opus}         & v14.10           &  \\
Tatoeba \citep{tiedemann2012opus}        & v2            &  \\
GNOME \citep{tiedemann2012opus}          & v1          &  \\
GlobalVoices  \citep{tiedemann2012opus}  & v2018q4         &  \\
KDE4  \citep{tiedemann2012opus}          & v2               &  \\
KDE-Doc  \citep{tiedemann2012opus}       & v1             &  \\
PHP \citep{tiedemann2012opus}            & v1           &  \\
Wikipedia \citep{Wo_k_2014}      & v1.0             &  \\
Wikimedia \citep{tiedemann2012opus}      & v20210402       &  \\
JRC \citep{tiedemann2012opus}            & v3.0           &  \\
DGT \citep{tiedemann2012opus}            & v2019              &  \\
EuroPat \citep{europat}        & v3              &  \\
EUbookshop \citep{tiedemann2012opus}     & v2                &  \\
EMEA \citep{tiedemann2012opus}           & v3                &  \\
EUConst \citep{tiedemann2012opus}        & v1               &  \\
tico-19 \citep{anastasopoulos-etal-2020-tico}        & v20201028        &  \\
ECB \citep{tiedemann2012opus}            & v1                 &  \\
Elitr-ECA \citep{williams2021elitr}      & v1            &  \\
MultiUN \citep{eisele-chen-2010-multiun}        & v1               &  \\
OpenOffice \citep{tiedemann2012opus}     & v3                 &  \\
Ada83 \citep{tiedemann2012opus}          & v1                &  \\
infopankki \citep{tiedemann2012opus}     & v1          &  \\
Scielo \citep{soares2018large}         & v1               &  \\
giga-fren \citep{tiedemann2012opus}      & v2               &  \\
UNPC \citep{ziemski-etal-2016-united}           & v1.0           & \\ \bottomrule
\end{tabular} 
\end{center}
\caption{Data sources from which we collect parallel data along with the datasets' version.}
\label{tab:parallel_data_sources}
\end{table}

\paragraph{\textbf{Code / Math Data}}
We collect code and mathematical data from The Stack \citep{Kocetkov2022TheStack}, the Algebraic-stack \citep{azerbayev2023llemma}, and Open-web-math \citep{paster2023openwebmath}.

For the 2\textsuperscript{nd} and 3\textsuperscript{rd} training phases we also incorporate the python-edu dataset \citep{benallal2024smollmcorpus} and the training sets of GSM8k \citep{cobbe2021gsm8k} and of Mathematics Aptitude Test of Heuristics \citep{hendrycksmath2021}. 

For the 3\textsuperscript{rd} phase, we also include synthetic data~(around 1.7 million samples) generated using the Qwen-2.5 models~\citep{qwen2025qwen25technicalreport, yang2024qwen25mathtechnicalreportmathematical}, which were used not only to rewrite questions, but also to generate answers to original data from MathInstruct~\citep{toshniwal2024openmathinstruct118millionmath, toshniwal2024openmathinstruct2acceleratingaimath} and MetaMathQA~\citep{yu2024metamathbootstrapmathematicalquestions}. These answers were then evaluated using techniques such as LLM-as-a-Judge~\citep{zheng2023judgingllmasajudgemtbenchchatbot}. Specifically, this involved generating answers using Qwen2.5-Math-7B, and subsequently evaluating them with Qwen2.5-Instruct as an LLM-as-a-Judge (filtering at a score of 9/10). Additionally, we also drawn samples from those datasets, and generated multiple-choice questions based on the original data, employing \texttt{gemma-2-9b-it}. The dataset was further augmented with samples from \texttt{SlimOrca}, which included original prompts and generations from \texttt{gemma-2-9b-it}, \texttt{gemma-2-27b-it}, \texttt{Llama-3.1-70B-Instruct}, and \texttt{Qwen2.5-32B-Instruct}; for these, \texttt{Qwen2.5-Instruct} provided judgements to ascertain the 'best-of-N' answer, with random sampling applied in cases of tied maximum ratings.

\paragraph{\textbf{Higher-quality Data}}
Regarding high-quality data, we use the Wikipedia \citep{wikidump} for all languages and the Arxiv \citep{clement2019}, Books \citep{Zhu_2015_ICCV}, and Apollo \citep{wang2024apollo} for English.

For the 2\textsuperscript{nd} and 3\textsuperscript{rd} training phases we add the Cosmopedia dataset (2\textsuperscript{nd} version; \citet{benallal2024smollmcorpus}). 
For the 3\textsuperscript{rd} we further include documents of Cosmopedia translated using Tower \citep{alves2024tower} to German, Spanish, French, Italian, Portuguese, Dutch, Chinese, and Russian. 

\subsection{Division Across Languages}
\begin{figure}[t]
    \centering
    \includegraphics[width=\textwidth]{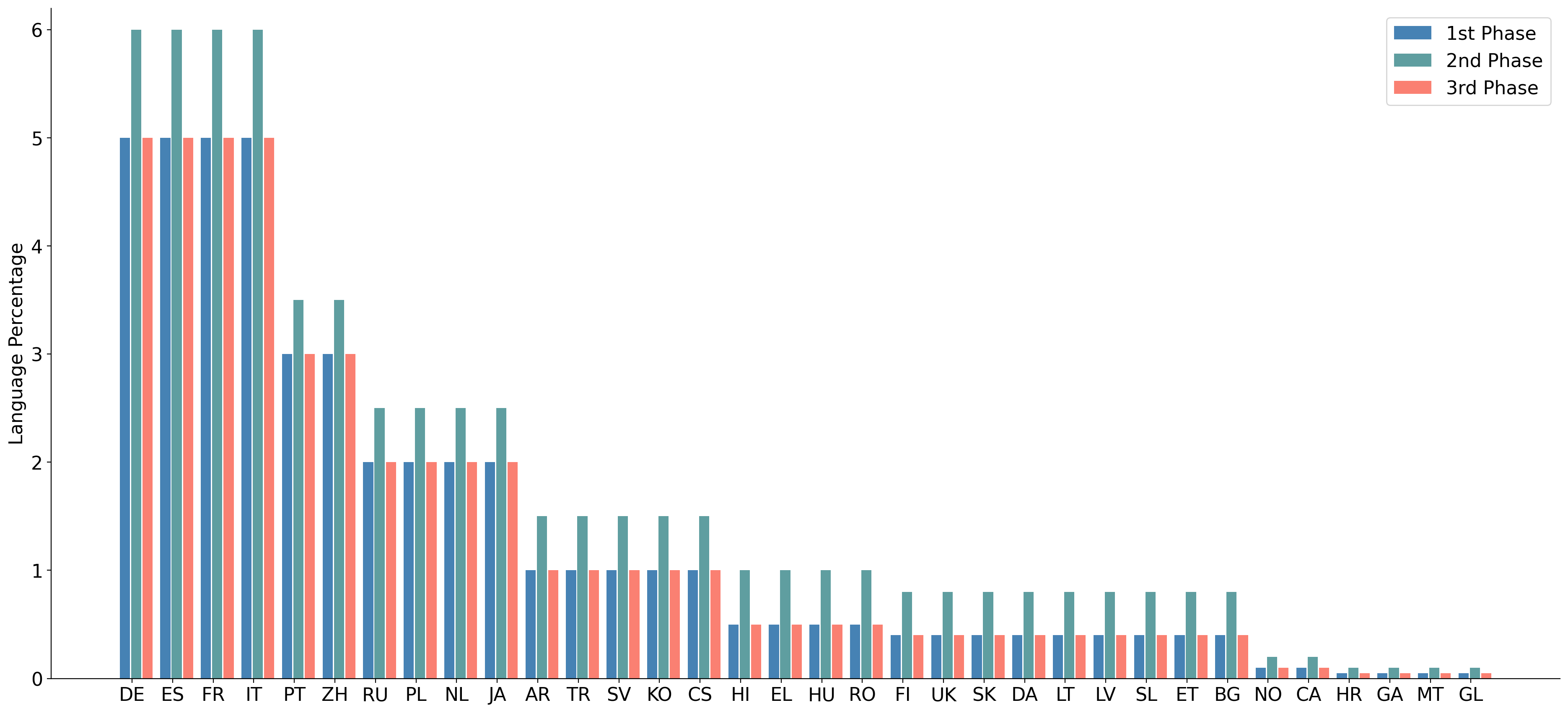}
    \caption{Percentage of the training corpus attributed to each language, excluding English and code / math data. English accounts to 50\% in the 1\textsuperscript{st} phase and  32.5\% during the 2\textsuperscript{nd} phase and 3\textsuperscript{rd} phases. 5\% of the corpus is left for datasets composed of code and math in the first phase, 7\% during the 2\textsuperscript{nd} phase and 23\% during the 3\textsuperscript{rd} phase.}
    \label{fig:language_percentage}
\end{figure}

The training corpus distribution evolves across phases to optimize multilingual capabilities and reasoning skills:
\begin{itemize}
    \item \textbf{1\textsuperscript{st} Phase}: During this phase, we designate 50\% for English, as both high-quality data and web data are predominantly in English, and include 5\% of code / math data. The remaining 45\% of the tokens are distributed among the other languages based on the amount of data obtained after the collection and filtering processes as shown by the blue bars on Fig. \ref{fig:language_percentage}. 
    \vspace{0.15cm}
    \item \textbf{2\textsuperscript{nd} Phase}: In order to increase EuroLLM's multilinguality, we decrease the English allocation to 32.5\% and distribute the surplus across the other languages. We also increase the code / math allocation to 7\%. See analysis in \S\ref{sec:2phase_experiments}.
    \vspace{0.15cm}
    \item \textbf{3\textsuperscript{rd} Phase}: Finally, to improve the model's reasoning abilities, in this last phase, we increase the code / math allocation to 23\%, by maintaining the English allocation but reverting the multilingual increase done for the 2\textsuperscript{nd} phase. See analysis in \S\ref{sec:3phase_experiments}. This is the main difference in terms of data and training to that of the EuroLLM-1.7B model~\citep{martins2024eurollm}.
\end{itemize}

Figure \ref{fig:language_percentage} shows the exact percentage attributed to each language in the three pre-training phases. These design decisions for data mixes were done with careful scaling laws that are thoroughly described in~\citet{martins2024eurollm}.

\subsubsection{Pre-training Progress}
We track the performance of EuroLLM-9B throughout the pre-training process along its three different phases.

\begin{figure}[t]
    \centering
    \includegraphics[width=4.73cm]{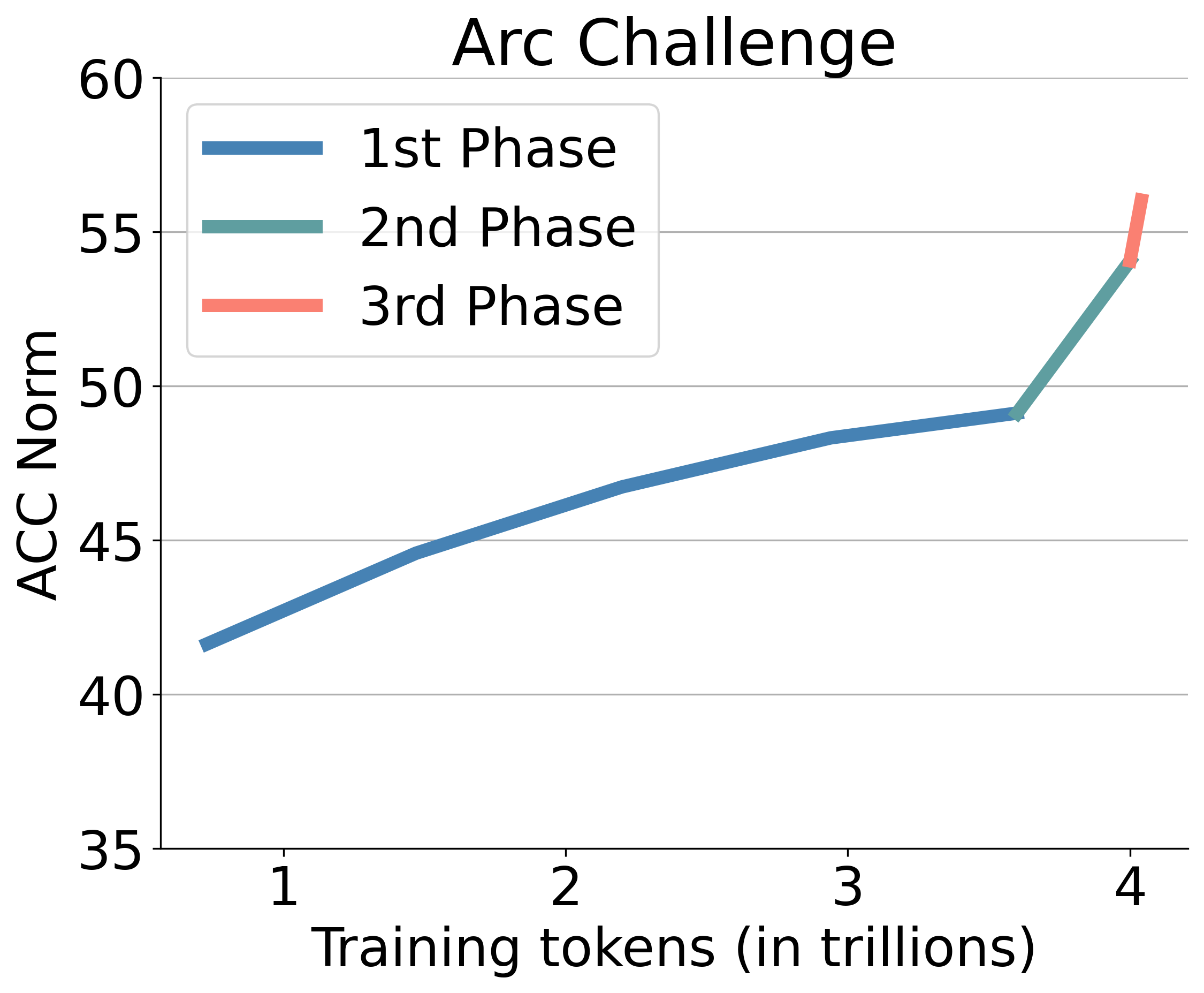}
    \includegraphics[width=4.535cm]{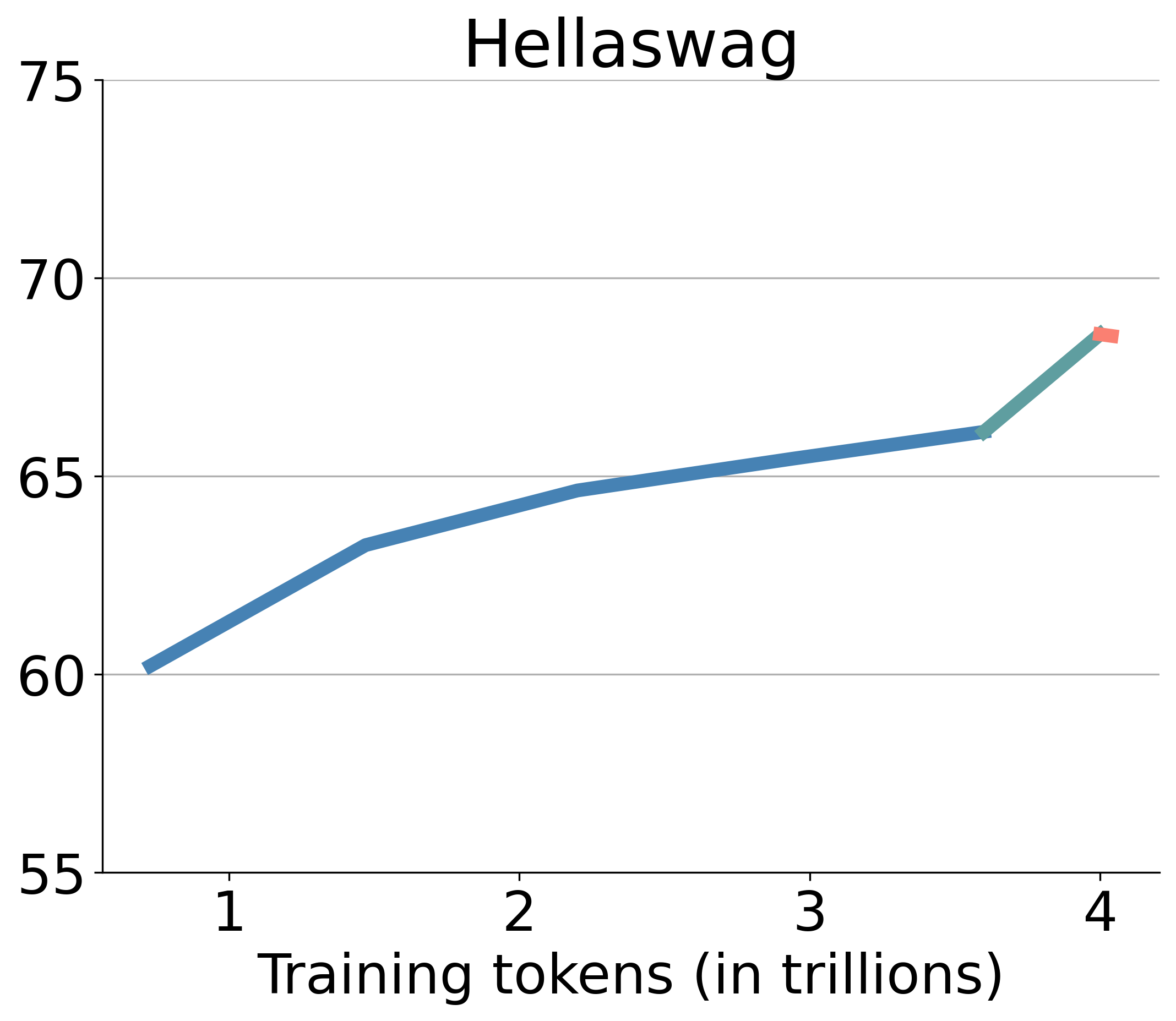}
    \includegraphics[width=4.535cm]{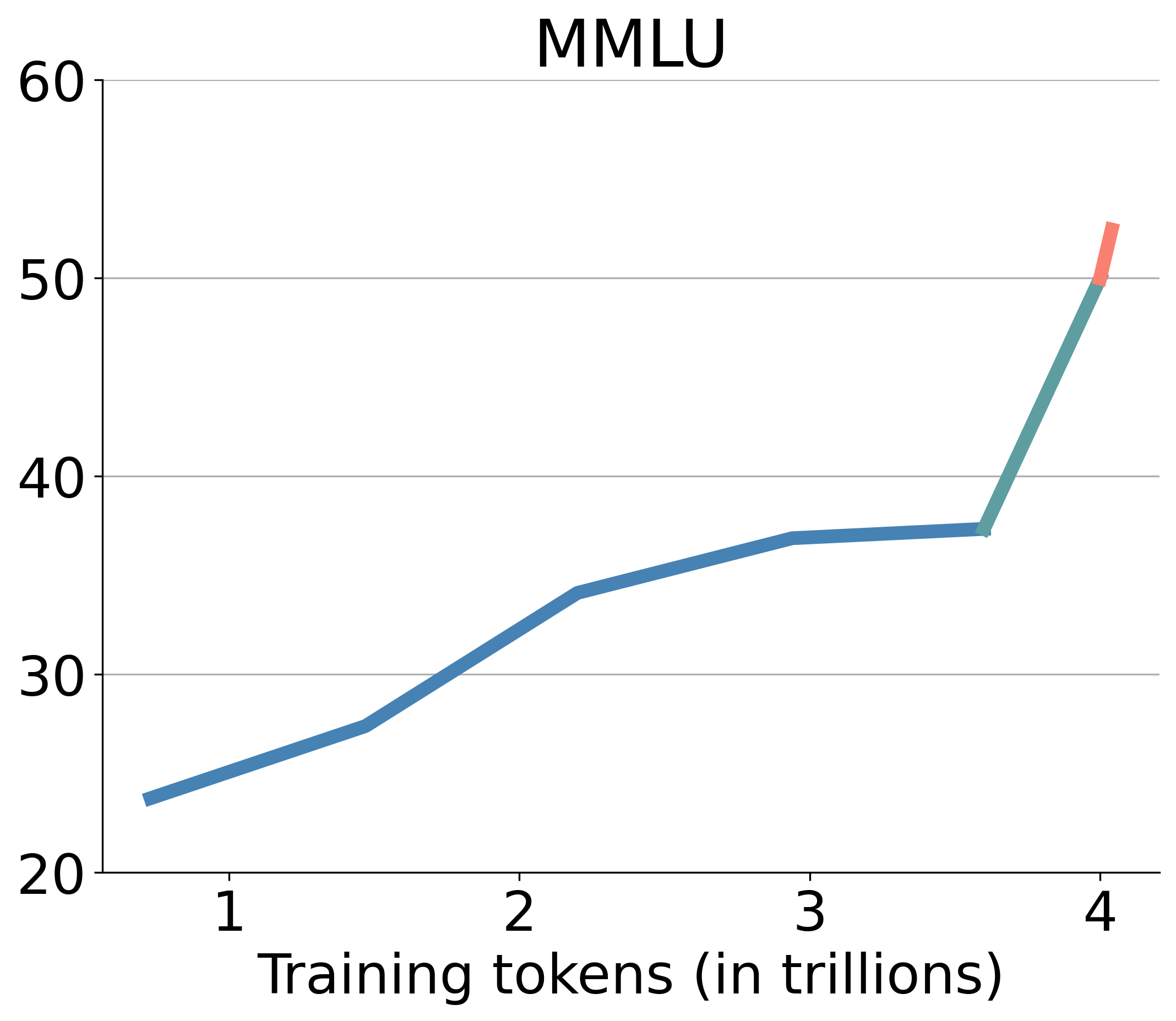}
    \caption{Results on Arc Challenge (left), Hellaswag (middle), and MMLU (right) throughout the pre-training process averaged across 11 EU languages.}
    \label{fig:results_analysis}
\end{figure}

The results, shown in Figure~\ref{fig:results_analysis}, demonstrate consistent improvement across all benchmarks during pre-training, with particularly notable gains in the second phase. During the third phase, there is an even steeper improvement on Arc Challenge and MMLU, but a slight decline on Hellaswag. We attribute this decrease to the increased proportion of code and math data in the corpus used for the third pre-training phase.

\section{Post Training}
\label{sec:post_training}
In this section, we describe the post-training process of EuroLLM-9B-Instruct, covering the post-training data—released as EuroBlocks (\S\ref{sec:post_training_data})—and the modeling decisions (\S\ref{sec:post_training_modeling}). We carried out the post-training using the Axolotl codebase\footnote{\url{https://docs.axolotl.ai/}}.

\subsection{Data}
\label{sec:post_training_data}

To enable EuroLLM-9B to follow natural language instructions, we constructed EuroBlocks, a multilingual dataset that combines both human-written and synthetic instruction-following conversations. The human-written portion draws from several publicly available sources, including Magpie \citep{xu2024magpie}\footnote{Magpie datasets are generated with several models with different licenses. We used only the data from Qwen 2, Llama 3 and Phi 3 models which have commercially permissive licenses.}, Aya \citep{singh2024aya}, lmsys-chat-1m \citep{zheng2023lmsyschat1m}, OpenMath-2 \citep{toshniwal2024openmath2}, and smol-magpie-ultra \citep{allal2024SmolLM2}.

To ensure data quality, we applied filtering based on complexity and readability scores. Prompts from OpenMath-2 and smol-magpie-ultra falling below a score of 4 on either dimension were removed, while low-scoring prompts from Magpie, Aya, and lmsys-chat-1m were downsampled rather than discarded entirely. We further filtered conversations using ArmoRM-v0.1~\citep{wang-etal-2024-interpretable}, removing responses with scores below 0.08---except in cases where low readability in the prompt skewed the score.\footnote{ArmoRM-v0.1 was found to be robust on multilingual data, with a Pearson correlation above 0.7 between English and translated examples. While reward models yield uncalibrated scores, the 0.08 threshold provided a good balance between quality and data retention.}

To broaden language coverage and support less-represented languages, we generated synthetic data. This involved prompting an LLM---either Llama 3 \citep{llama3modelcard} or an earlier EuroLLM checkpoint---with a monolingual document, a target language, and a category, asking it to create an instruction relevant to the document (see prompt in Fig. \ref{fig:synthethic-data-prompt}). The same model then produced an answer in a RAG-style setup, using both the document and the generated instruction (see prompt in Fig. \ref{fig:synthethic-data-answer}). Additionally, we synthesized further supervised fine-tuning (SFT) data by translating prompt–answer pairs,\footnote{Translations were produced using Tower v2~\citep{rei-etal-2024-tower} or earlier EuroLLM-9B models.} and incorporating high-quality examples from multilingual translation benchmarks such as NTREX-128 \citep{federmann-etal-2022-ntrex}, \textsc{Flores-200-dev} \citep{nllb2022}, WMT-21 \citep{farhad2021findings}, and WMT-22 \citep{kocmi2022findings}, leaving WMT-23 and later editions for evaluation purposes.

Altogether, we collected approximately 4.5 million instructions. After applying filtering and deduplication, the final EuroBlocks dataset contains 1.95 million high-quality examples. To support further research and development of European-centric LLMs, we publicly release the synthetic portion of the dataset: \href{https://huggingface.co/datasets/utter-project/EuroBlocks-SFT-Synthetic-1124}{utter-project/EuroBlocks-SFT-Synthetic-1124}.

\subsection{Modeling}
\label{sec:post_training_modeling}
We fine-tune EuroLLM-9B on EuroBlocks to create an instruction-following conversational model, EuroLLM-9B-Instruct. We use the standard cross-entropy loss, enabling \texttt{bfloat16} mixed precision and packing. The loss is calculated only on target tokens, with the loss on prompt tokens being masked. The model is trained for three epochs using a learning rate of $7 \times 10^{-6}$.

\section{Evaluation}
\label{sec:results}

We evaluate EuroLLM-9B's performance by comparing it against publicly available models (listed in \S\ref{sec:baselines}) and analyzing its progression throughout pre-training (\S\ref{sec:analysis}). Our evaluation encompasses average performance across EU official languages (\S\ref{sec:eu_languages}), per-language results for EU languages (\S\ref{sec:eu_lang_by_lang}), and results for additional supported languages by EuroLLM-9B (\S\ref{sec:others_lang_by_lang}).

\subsection{Evaluation Settings}
\label{sec:eval_settings}
Our assessment framework encompasses two main categories: general benchmarks to evaluate world knowledge acquisition, and machine translation to assess multilingual understanding and generation.

Regarding English general benchmarks, we evaluate the pre-trained and post-trained versions of each LLM (when available) on:
\begin{itemize}
    \item Arc-Easy and Arc-Challenge \citep{allenai:arc}: multiple-choice question-answering dataset, containing questions from science exams from grade 3 to grade 9. The Challenge partition contains the most difficult questions.
    \item Hellaswag \citep{zellers2019hellaswag}: multiple-choice commonsense inference test which requires the understanding not just of the words in the sentence, but also of the underlying meaning and context.
    \item MMLU \citep{hendryckstest2021}: multiple-choice questions from various branches of knowledge: humanities, social sciences, hard sciences, and other areas 
    \item MMLU-PRO \citep{wang2024mmlu}: refined and more challenging version of the MMLU dataset.
    \item MUSR \citep{spraguemusr}: multiple-choice complex problems with around 1,000 words in length generated algorithmically. These problems, which include murder mysteries, object placement questions, and team allocation problems require models to reason with long-range context. 
    \item TruthfulQA \citep{lin-etal-2022-truthfulqa}: multiple-choice questions designed to evaluate the model's ability to identify true statements. It contains 817 questions from 38 different categories, including health, law, finance and politics.
    \item IFEval \citep{kovalevskyi2024ifeval}: set of prompts that test a model’s ability to follow explicit instructions, such as ``include keyword x'' or ``use format y''.
\end{itemize}

To ensure comprehensive multilingual evaluation, in the final evaluations we employed translated versions of six benchmarks: Arc (both Easy and Challenge partitions), Hellaswag, MMLU, TruthfulQA, MMLU-Pro, and MUSR. 
For European languages, we utilized the EU20-Benchmarks\footnote{\href{https://huggingface.co/collections/openGPT-X/eu20-benchmarks-67093b13db8ff192dc39312d}{https://huggingface.co/collections/openGPT-X/eu20-benchmarks-67093b13db8ff192dc39312d}}  \citep{eu20}, which contains translations of the first four benchmarks across all target European languages except Irish, Maltese, and Croatian. 
For non-European languages, we used the translated versions of Arc-challenge, Hellaswag, and MMLU from the Okapi benchmark collection \citep{lai2023okapi}.
For MMLU-Pro and MUSR benchmarks, no existing translations were available, so we created our own translations using Tower \citep{alves2024tower}, covering a subset of the languages supported in EuroLLM. 
And finally, for English we also used IF-Eval in addition to the other test sets.\footnote{For reproducibility, we release our evaluation code, detailled parameter settings and result files: \href{https://github.com/utter-project/eurollm-evaluation}{https://github.com/utter-project/eurollm-evaluation}.}

Regarding evaluation, we follow the \href{https://huggingface.co/spaces/open-llm-leaderboard/open_llm_leaderboard#/}{Open LLM Leaderboard} methodology for MMLU-PRO, MUSR, and IFEval, normalizing scores between random baseline and maximum possible score (see the leaderboard \href{https://huggingface.co/spaces/open-llm-leaderboard/blog}{blog post} for details). For the remaining benchmarks, we adhere to the \href{https://huggingface.co/docs/leaderboards/open_llm_leaderboard/archive}{Open LLM Leaderboard v1} specifications.

For machine translation evaluation, we use the WMT24++ dataset \citep{deutsch2025wmt24expandinglanguagecoverage} and report results using the COMET-22 metric \citep{rei2022comet} in both directions: from and into English. WMT24++ extends the official WMT24 dataset by providing new post-edited references for 8 out of 9 original language pairs, as well as new human-written references and post-edits for 46 additional languages and dialects, covering a total of 55 languages.

We also provide the Borda count \citep{irurozki2022best}, which corresponds to the average ranking of the models.

\subsection{Baselines}
\label{sec:baselines}
Our evaluation includes a comprehensive comparison with publicly available LLMs of the same size range, categorized into European-made and non-European-made models, considering both pre-trained and post-trained versions. The full list of models can be found in Table~\ref{table:baselines}.

\begin{table}
\footnotesize
\setlength{\tabcolsep}{0.55ex}
\begin{tabular}{llccc}
\toprule
\textbf{Pre-trained} & \textbf{Post-trained} & \textbf{Technical Report} &\textbf{European} & \textbf{EU Lang. Supp.}  \\
\midrule
\href{https://huggingface.co/google/gemma-2-9b}{Gemma-2-9B} & \href{https://huggingface.co/google/gemma-2-9b-it}{Gemma-2-9B-IT} & \citet{team2024gemma2} & No & ----- \\
\href{https://huggingface.co/meta-llama/Llama-3.1-8B}{LLaMa-3.1-8B} & \href{https://huggingface.co/meta-llama/Llama-3.1-8B-Instruct}{LLaMa-3.1-8B-IT} & \citet{dubey2024llama} & No & ----- \\
\href{https://huggingface.co/ibm-granite/granite-3.1-8b-base}{Granite-3-8B} & \href{https://huggingface.co/ibm-granite/granite-3.1-8b-instruct}{Granite-3-8B-IT} & \citet{granite2024granite} & No & No \\
\href{https://huggingface.co/Qwen/Qwen2.5-7B}{Qwen-2.5-7B} & \href{https://huggingface.co/Qwen/Qwen2.5-7B-Instruct}{Qwen-2.5-7B-IT} & \citet{qwen2025qwen25technicalreport} & No & No \\
\href{https://huggingface.co/allenai/OLMo-2-1124-7B}{OLMo-2-7B} & \href{https://huggingface.co/allenai/OLMo-2-1124-7B-Instruct}{OLMo-2-7B-IT} & \citet{olmo20242} & No & No \\
\href{https://huggingface.co/CohereForAI/aya-23-8B}{Aya-23-8B} & \href{https://huggingface.co/CohereForAI/aya-expanse-8b}{Aya-Expanse-8B} & \citet{singh2024aya,dang2024aya} & No & No \\
\href{https://huggingface.co/mistralai/Mistral-7B-v0.3}{Mistral-7B} & \href{https://huggingface.co/mistralai/Mistral-7B-Instruct-v0.3}{Mistral-7B-IT} & \citet{jiang2023mistral} & Yes & No \\
Not available & \href{https://huggingface.co/mistralai/Ministral-8B-Instruct-2410}{Ministral-8B-IT} & ---- & Yes & No \\
\href{https://huggingface.co/occiglot/occiglot-7b-eu5}{Occiglot-7B-eu5} & \href{https://huggingface.co/occiglot/occiglot-7b-eu5-instruct}{Occiglot-7B-eu5-IT} & ---- & Yes & No \\
\href{https://huggingface.co/BSC-LT/salamandra-7b}{Salamandra-7B} & \href{hhttps://huggingface.co/BSC-LT/salamandra-7b-instruct}{Salamandra-7B-IT} & \cite{gonzalezagirre2025salamandratechnicalreport} & Yes & Yes \\
Not available & \href{https://huggingface.co/Aleph-Alpha/Pharia-1-LLM-7B-control-hf}{Pharia-1-LLM-7B-C} & ---- & Yes & No \\
Not available & \href{https://huggingface.co/openGPT-X/Teuken-7B-instruct-research-v0.4}{Teuken-7B-IT-R-v0.4} & \citet{ali2024teuken} & Yes & Yes \\
Not available & \href{https://huggingface.co/openGPT-X/Teuken-7B-instruct-commercial-v0.4}{Teuken-7B-IT-C-v0.4} & \citet{ali2024teuken} & Yes & No \\
\bottomrule
\end{tabular}
\caption{List of pre-trained and post-trained LLMs which we compare with EuroLLM-9B.}
\label{table:baselines}
\end{table}

\subsection{Results}
\label{sec:eu_languages}

The analyses of pre-trained (Table \ref{table:eu_results_pre}) and post-trained models (Table \ref{table:eu_results_post}) across EU languages demonstrate EuroLLM-9B's strong performance. Both the base model and its post-trained variant (EuroLLM-9B-IT) emerge as the top performers among European-made models, achieving superior results across most benchmarks as reflected in their lowest Borda count scores. Furthermore, EuroLLM-9B shows performance comparable to Gemma-2-9B while outperforming the remaining non-European-made LLMs on the majority of the evaluated tasks.
Comparing model performance on the machine translation task reveals that EuroLLM-9B-IT achieves the best results among all European and non-European models, in both translation directions (from and into English), with more than three points of difference with the second best model Gemma-2-9B-IT, which is a very strong result. 



\begin{table}
\resizebox{\textwidth}{!}{%
\begin{tabular}{lccccccc}
\toprule
Pre-trained & \textbf{Arc} & \textbf{Hellaswag} & \textbf{MMLU} & \multicolumn{1}{l}{\textbf{TruthfulQA}} & \textbf{MMLU-pro} & \textbf{MUSR} & \textbf{Borda C $\downarrow$} \\
 & (25-shot) & (10-shot) & (5-shot) & (0-shot) & (5-shot) & (0-shot) &  \\ \midrule
\textit{\textbf{Non-European}} &  &  &  & \multicolumn{1}{l}{} &  &  &  \\ \midrule
Gemma-2-9B & \textbf{67.89} & \textbf{67.73} & \textbf{66.19} & 50.63 & 29.75 & \textbf{9.70} & \textbf{1.3} \\
LLaMa-3.1-8B & 55.46 & 58.86 & 55.54 & 49.49 & 19.94 & 5.44 &  3.0 \\
Granite-3-8B & 47.42 & 51.73 & 47.10 & 49.34 & 20.38 & 7.07 &  4.0 \\
Qwen-2.5-7B & 50.68 & 52.17 & 62.44 & \textbf{54.06} & \textbf{31.63} & 8.04 & 2.2 \\
OLMo-2-7B & 38.25 & 42.23 & 41.32 & 45.24 & 13.91 & 4.53 & 5.8 \\
Aya-23-8B & 47.53 & 53.48 & 45.44 & 47.64 & 14.04 & 3.64 & 4.7 \\ \midrule
\textit{\textbf{European}} &  &  &  & \multicolumn{1}{l}{} &  &  &  \\ \midrule
Mistral-7B & 51.31 & 53.38 & 50.09 & 47.15 & 17.36 & 8.69 & 2.3 \\
Occiglot-7B-eu5 & 46.90 & 51.12 & 42.52 & 44.81 & 11.98 & 3.83 & 3.7 \\
Salamandra-7B & 61.15 & 64.73 & 42.75 & 46.06 & 5.25 & 2.63 & 3.0 \\
\textbf{EuroLLM-9B} & \textbf{66.48} & \textbf{67.00} & \textbf{55.68} & \textbf{51.84} & \textbf{17.60} & \textbf{10.97} & \textbf{1.0} \\ \bottomrule
\end{tabular}%
}
\caption{Comparison of the pre-trained versions of open-weight LLMs on multilingual benchmarks, averaged across EU official languages. 
For Arc, Hellaswag, MMLU, and TruthfulQA we use EU20 benchmark \citep{eu20}. 
For MMLU-Pro and MUSR we translate the English version with Tower \citep{alves2024tower} to 7 EU languages (German, French, Spanish, Portuguese, Italian, Dutch, and Czech).
Scores of MMLU-PRO, and MUSR are normalized between random baseline and maximum possible score, following the methodology used in \textit{Open LLM Leaderboard}.}
\label{table:eu_results_pre}
\end{table}

\begin{table}[]
\resizebox{\textwidth}{!}{%
\begin{tabular}{lccccccccc}
\toprule
Post-trained & \textbf{Arc} & \textbf{Hellaswag} & \textbf{MMLU} & \textbf{TruthfulQA} & \textbf{MMLU-pro} & \textbf{MUSR} & \textbf{WMT24++} & \textbf{WMT24++} & \textbf{Borda C $\downarrow$} \\
 & \multicolumn{1}{l}{} & \multicolumn{1}{l}{} & \multicolumn{1}{l}{} & \multicolumn{1}{l}{} & \multicolumn{1}{l}{} & \multicolumn{1}{l}{} & en→xx & xx→en & \multicolumn{1}{l}{} \\
 & (25-shot) & (10-shot) & (5-shot) & (0-shot) & (5-shot) & (0-shot) & (0-shot) & (0-shot) &  \\ \midrule
\textit{\textbf{Non-European}} &  &  &  &  &  &  &  &  &  \\ \midrule
Gemma-2-9B-IT & \textbf{64.60} & \textbf{64.28} & \textbf{65.13} & \textbf{60.65} & 27.42 & \textbf{8.38} & \textbf{80.47} & \textbf{80.39} & \textbf{1.1} \\
LLaMa-3.1-8B-IT & 51.39 & 56.92 & 57.07 & 55.05 & 24.22 & 4.01 & 77.43 & 77.70 & 3.0 \\
Granite-3-8B-IT & 46.98 & 52.86 & 49.36 & 56.04 & 20.10 & 7.90 & 66.59 & 67.24 & 4.1 \\
Qwen-2.5-7B-IT & 47.91 & 51.61 & 62.27 & 57.88 & \textbf{29.68} & 7.62 & 69.32 & 69.54 & 3.0 \\
OLMo-2-7B-IT & 39.00 & 43.30 & 41.86 & 48.57 & 12.38 & 4.02 & 62.43 & 63.20 & 5.9 \\
Aya-Expanse-8B & 47.78 & 54.94 & 51.33 & 53.57 & 19.77 & 5.52 & 72.02 & 73.90 & 3.9 \\ \midrule
\textit{\textbf{European}} &  &  &  &  &  &  &  &  &  \\ \midrule
Mistral-7B-IT & 52.63 & 53.40 & 48.29 & \textbf{58.01} & \textbf{18.19} & 6.94 & 70.08 & 70.01 & 4.0 \\
Ministral-8B-IT & 51.71 & 55.36 & 51.22 & 52.53 & 17.41 & 6.17 & 73.52 & 73.77 & 4.1 \\
Occiglot-7B-eu5-IT & 42.39 & 49.52 & 39.75 & 48.10 & 11.77 & 4.17 & 61.14 & 59.59 & 6.4 \\
Salamandra-7B-IT & 55.16 & 63.46 & 47.30 & 51.15 & 7.01 & 7.17 & 81.38 & 78.21 & 3.8 \\
Pharia-1-LLM-7B-C & 38.58 & 43.13 & 34.68 & 45.80 & 10.10 & \textbf{9.83} & 51.71 & 49.44 & 6.8 \\
Teuken-7B-IT-R-v0.4 & 55.42 & 60.96 & 37.65 & 54.75 & 9.29 & 2.25 & 75.19 & 70.50 & 4.8 \\
Teuken-7B-IT-C-v0.4 & 53.77 & 60.25 & 40.05 & 52.68 & 9.79 & 2.94 & 76.44 & 71.65 & 4.6 \\
\textbf{EuroLLM-9B-IT} & \textbf{60.67} & \textbf{64.94} & \textbf{55.37} & 53.99 & 17.04 & 9.02 & \textbf{84.19} & \textbf{83.94} & \textbf{1.6} \\ \bottomrule
\end{tabular}%
}
\caption{Comparison of the post-trained versions of open-weight LLMs on multilingual benchmarks, averaged across EU official languages. For WMT24++ we average the Comet-22 scores on all 21 language pairs which include English as source (for en→xx) or as the target language (for xx→en).
Scores of MMLU-PRO, and MUSR are normalized between random baseline and maximum possible score, following the methodology used in \textit{Open LLM Leaderboard}.}
\label{table:eu_results_post}
\end{table}

The results of the pre-trained and post-trained models for each language are presented in Appendix \ref{sec:results_per_lang}. 
As we can see in Tables \ref{tab:pretrained_bg}--\ref{tab:posttrained_sv}, in 13 languages, our pre-trained model outperforms the other European open-weight models in all the tasks. In the case of post-trained models, in 14 languages, EuroLLM-9B-IT outperforms all the other open-weight models in all the tasks, except TruthfulQA. For this specific task, it ranks fourth on average.
Our initial investigation suggests that it is mainly due to the challenging characteristics of this benchmark. It contains questions that some humans would answer falsely due to a false belief or misconception \citep{lin-etal-2022-truthfulqa}.
Finally, in the machine translation task, EuroLLM-9B-IT outperforms all other European models across all language pairs and translation directions, with the sole exception of Greek→English.

\section{Analysis and Discussion}
\subsection{Pre-training Analysis}
\label{sec:analysis}

In this section, we analyse the pre-training process averaging scores across 11 EU languages from the Okapi benchmark: English, German, Spanish, French, Italian, Portuguese, Dutch, Swedish, Hungarian, Romanian, and Danish. 



\subsubsection{Second Phase Data Mixtures}
\label{sec:2phase_experiments}

To determine the optimal data mixture for the second training phase, we perform several experiments using EuroLLM-1.7B \citep{martins2024eurollm} with reduced datasets of 80 billion tokens instead of the full 400 billion. We evaluate three distinct data mixtures, each maintaining an increased percentage of higher-quality data at 34\% while varying other components:
\begin{enumerate}
    \item English content reduced to 48\% with code/math data increased to 7\%.
    \item English content reduced to 40\% with code/math data increased to 15\%.
    \item English content reduced to 32.5\% with code/math data at 7\%, redistributing the remaining percentage across the other languages.
\end{enumerate}
\begin{figure}[h]
    \centering
    \includegraphics[width=4.73cm]{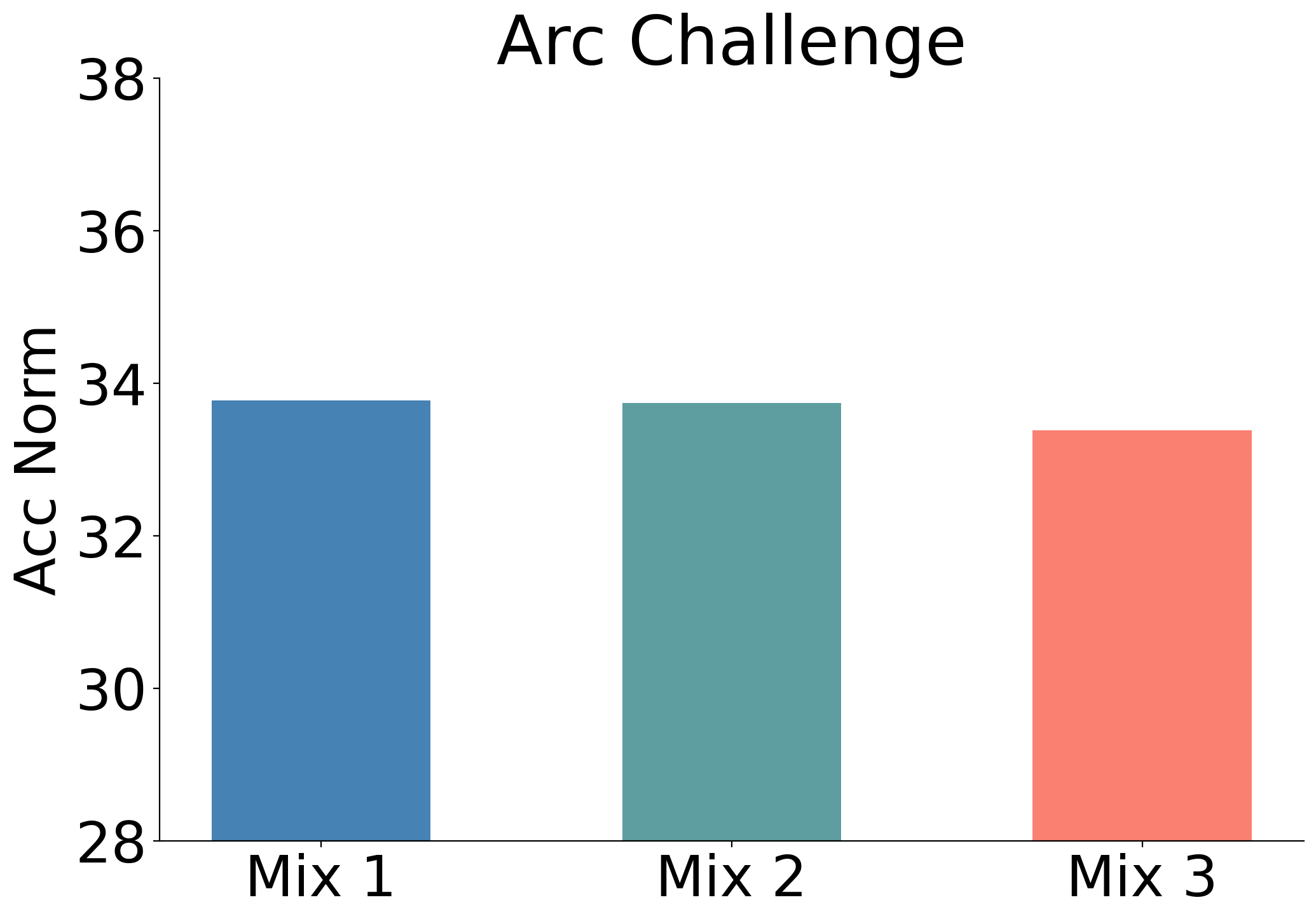}
    \includegraphics[width=4.535cm]{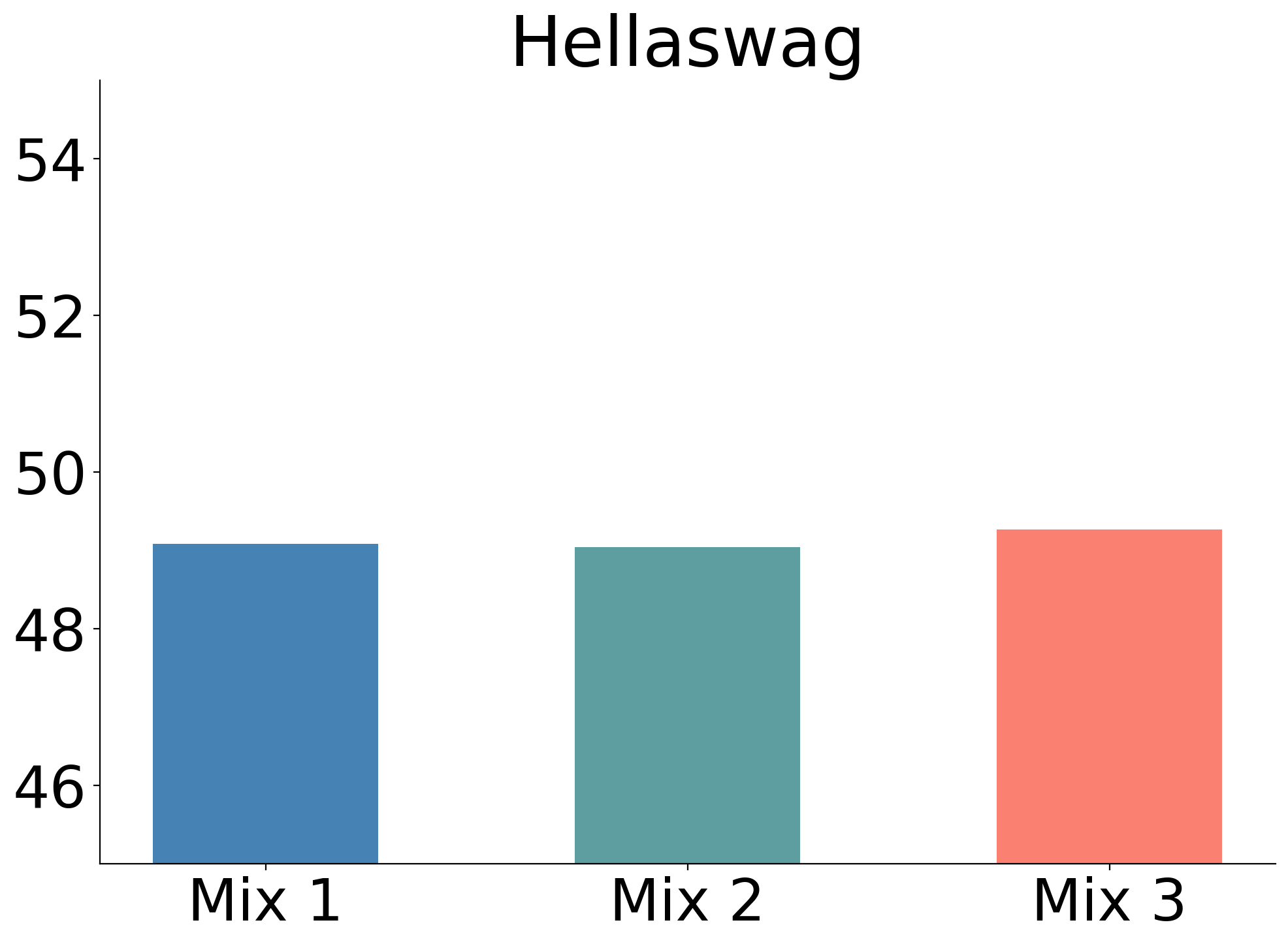}
    \includegraphics[width=4.535cm]{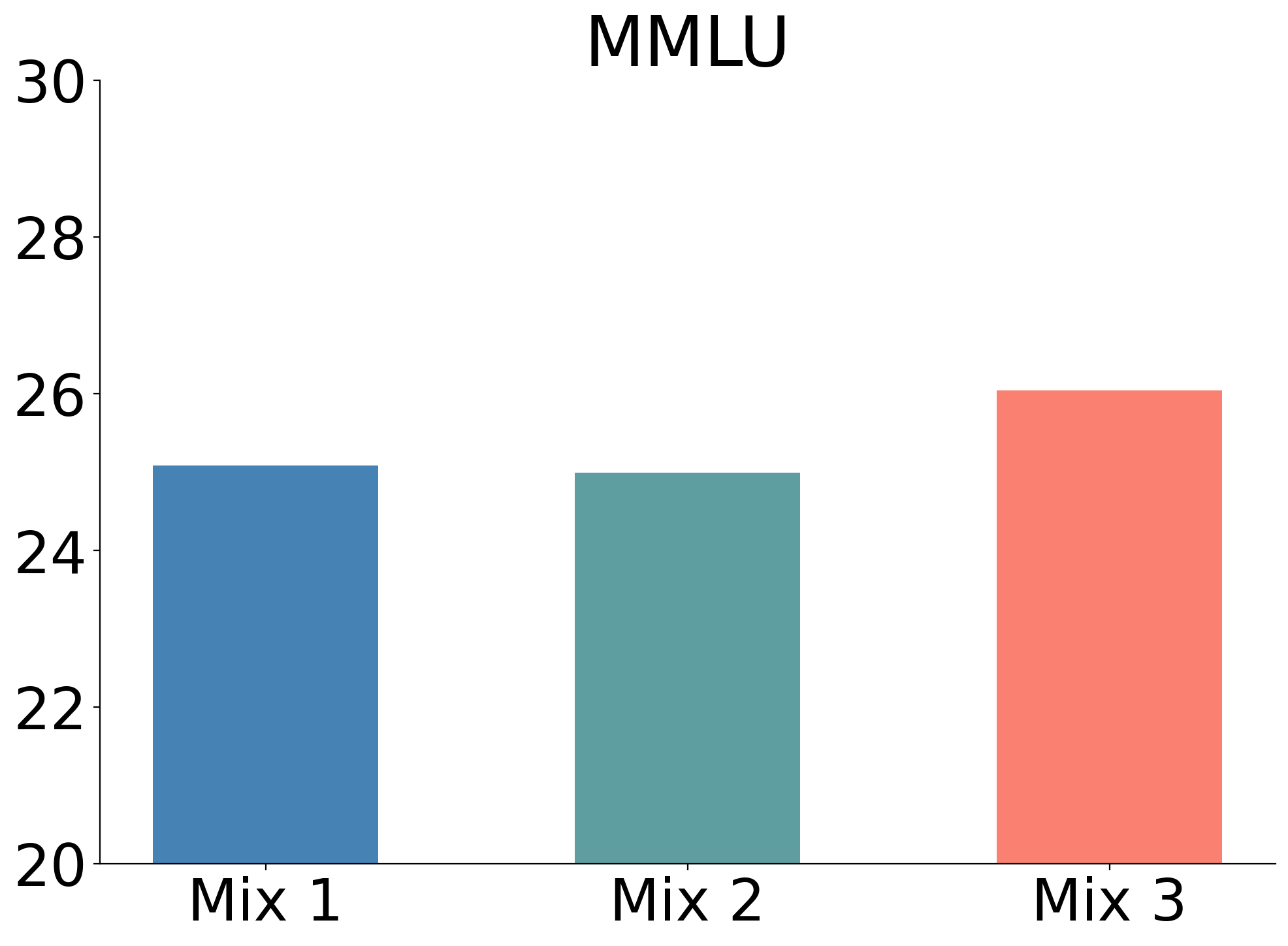}
    \caption{Results on Arc Challenge (left), Hellaswag (middle), and MMLU (right) with different 2\textsuperscript{nd} phase data mixes, averaged across 11 EU languages.}
    \label{fig:results_analysis_2nd}
\end{figure}
The experimental results, shown in Figure~\ref{fig:results_analysis_2nd}, demonstrate that the third data mixture, while showing slightly lower performance on Arc-Challenge, achieved superior results on Hellaswag and more predominantly on MMLU. Based on this performance profile, we select this data mixture for the second training phase.

\subsubsection{Third Phase Data Mixtures}
\label{sec:3phase_experiments}


Then, to decide what data mixture to use in the third pre-training phase, we experiment performing it on EuroLLM-9B with three data mixtures: 
\begin{enumerate}
    \item Code/math data increased to 9.5\% with English content reduced to 30\%.
    \item Code/math data increased to 23\% with proportional reductions across non-English languages.
    \item Code/math data increased to 23\%, reduced proportions for non-English languages, and parallel data decreased to 2\% with corresponding increases in other data sources.
\end{enumerate}
\begin{figure}[h]
    \centering
    \includegraphics[width=4.73cm]{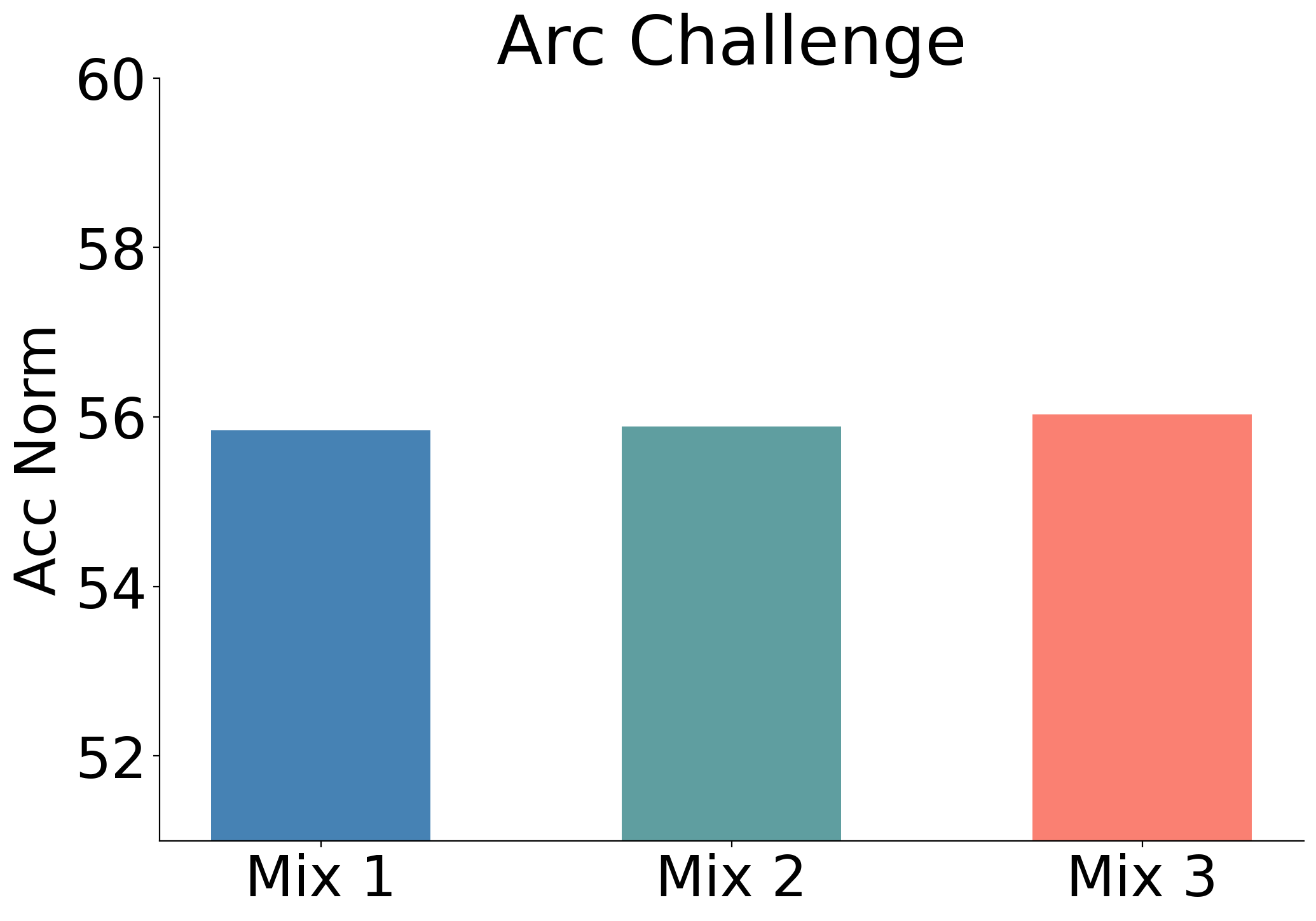}
    \includegraphics[width=4.535cm]{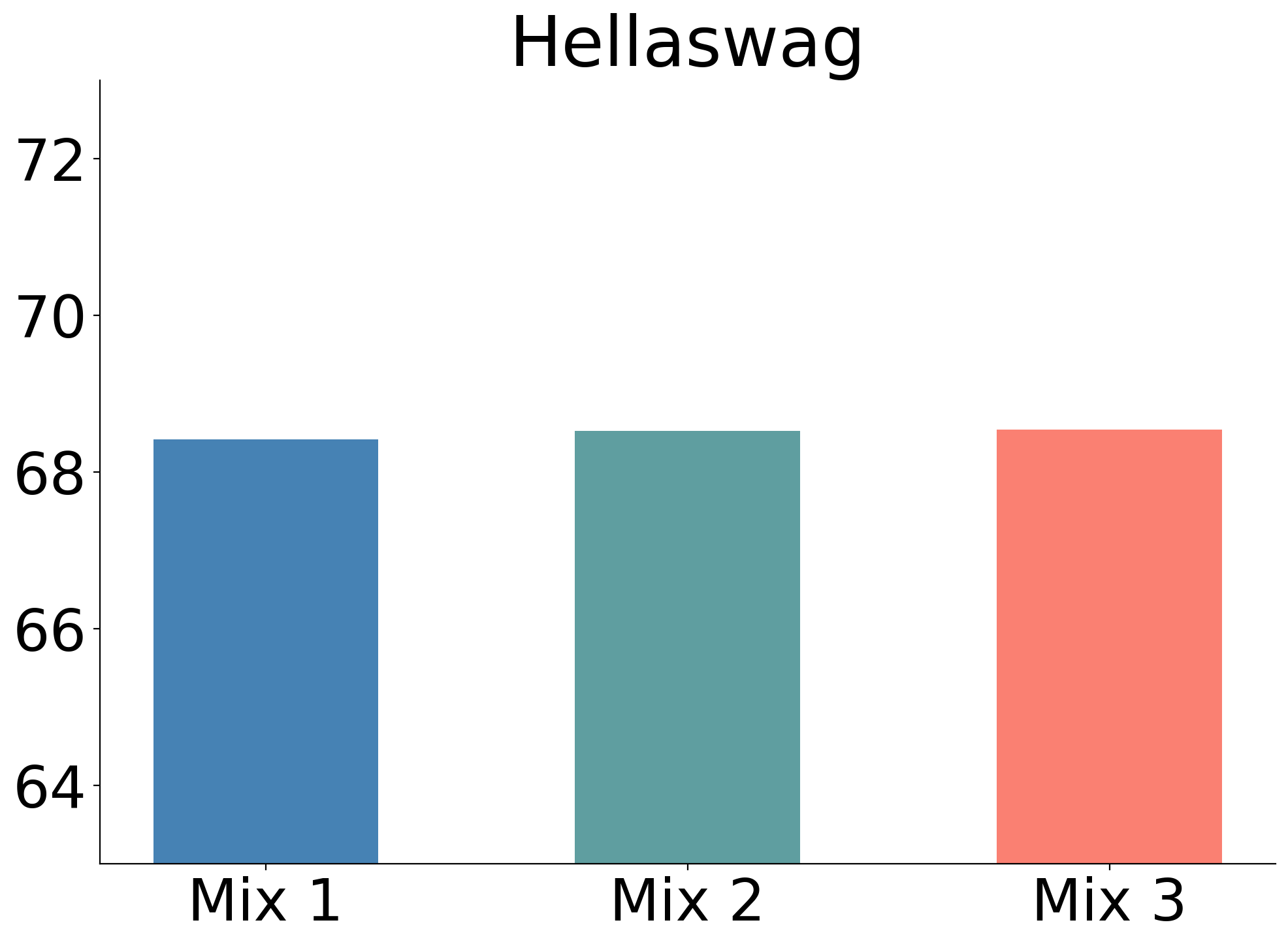}
    \includegraphics[width=4.535cm]{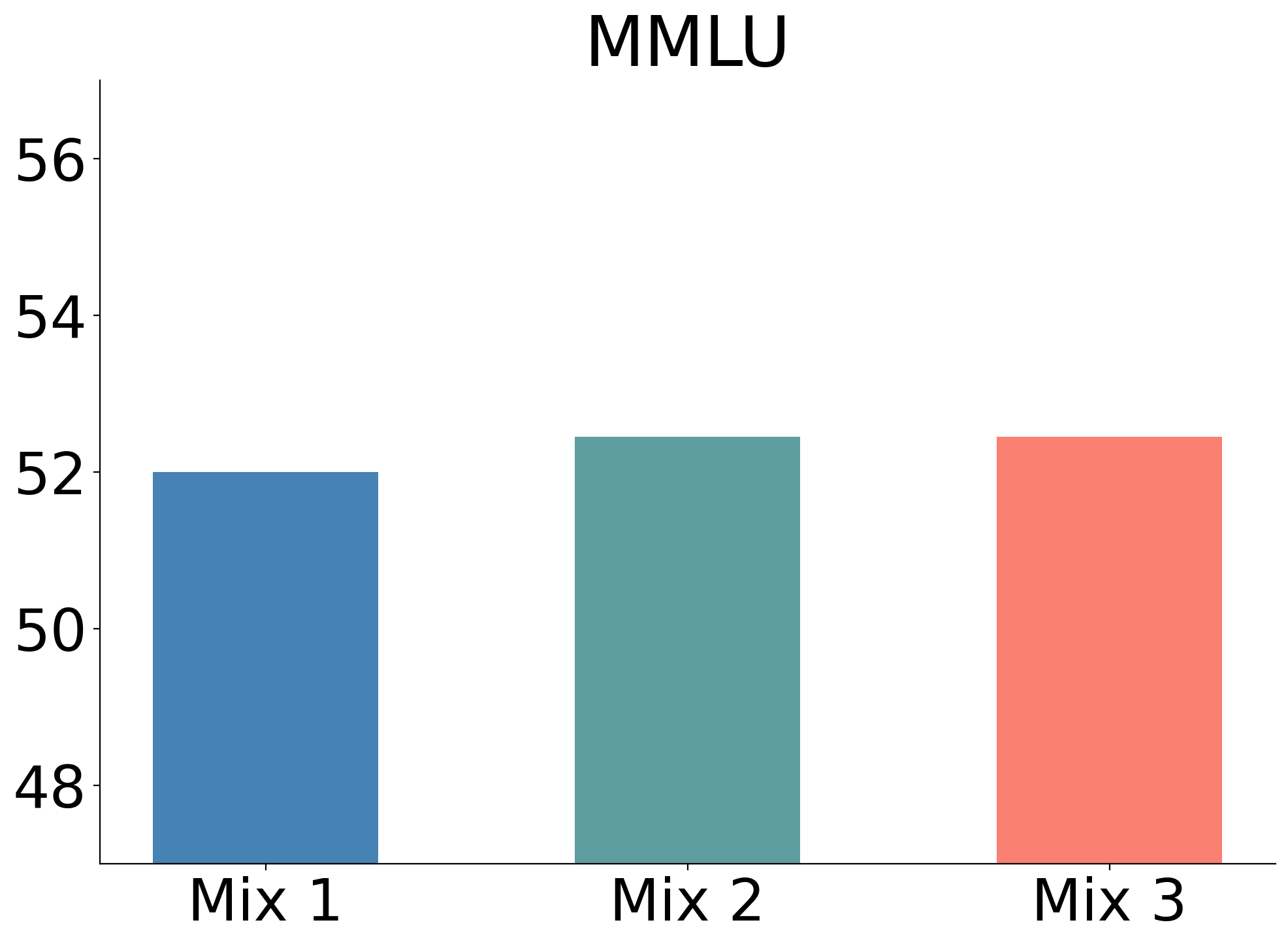}
    \caption{Results on Arc Challenge (left), Hellaswag (middle), and MMLU (right) with different 3\textsuperscript{rd} phase data mixes, averaged across 11 EU languages.}
    \label{fig:results_analysis_3rd}
\end{figure}

The experimental results, presented in Figure~\ref{fig:results_analysis_3rd}, demonstrate that the third configuration achieved equal or slightly superior performance on the three benchmarks. Based on these results, we select this mixture for the 3\textsuperscript{rd} pre-training phase.

\section{Conclusions}
\label{sec:conclusions}

We have presented EuroLLM-9B, detailing the comprehensive development process from data collection and filtering to the creation of our multilingual tokenizer and the pre-training and post-training processes. The release of both EuroLLM-9B and its instruction-tuned variant, EuroLLM-9B-Instruct, is accompanied by extensive performance evaluations on multilingual general benchmarks and machine translation tasks. In addition to the models, we also release EuroFilter, our multilingual data filtering model, and EuroBlocks-Synthetic, a synthetic instruction dataset designed to improve post-training coverage across European languages.

Looking ahead, we will continue to focus on developing larger-scale multilingual language models specifically designed for European languages and use cases.

\subsubsection*{Acknowledgments}
Part of this work was supported by the EU’s Horizon Europe Research and Innovation Actions (UTTER, contract 101070631), by the project DECOLLAGE (ERC-2022-CoG
101088763), and by the Portuguese Recovery and Resilience Plan through project C645008882-00000055 (Center for Responsible AI). We thank EuroHPC for the HPC resources used to support this work through grant EHPC-EXT-2023E01-042 and grant EHPC-AI-2024A01-085.

\newpage
\bibliography{iclr2025_conference}
\bibliographystyle{iclr2025_conference}

\newpage
\appendix
\section{Appendix}
\label{sec:results_per_lang}
Here we present the results of the pre-trained and post-trained models for each language separately.
\subsection{European Languages}
\label{sec:eu_lang_by_lang}
\subsubsection{Bulgarian (bg)}

\begin{table}[H]
\resizebox{\textwidth}{!}{%
\begin{tabular}{lccccc}
\toprule
Pre-trained & \textbf{Arc} & \textbf{Hellaswag} & \textbf{MMLU} & \textbf{TruthfulQA} & \textbf{Borda C}  $\downarrow$ \\
 & (25-shot) & (10-shot) & (5-shot) & (0-shot) &  \\ \midrule
\textit{\textbf{Non-European}} &  &  &  &  &  \\ \midrule

Gemma-2-9B & \textbf{67.87} & \textbf{68.32} & \textbf{64.34} & 54.37 & \textbf{1.2} \\

LLaMa-3.1-8B & 54.12 & 55.71 & 51.46 & 52.01 & 2.8 \\

Granite-3-8B & 42.17 & 44.64 & 33.08 & 52.45 & 4.0 \\

Qwen-2.5-7B & 47.95 & 48.95 & 61.22 & \textbf{55.84} & 2.2 \\

OLMo-2-7B & 33.67 & 37.79 & 29.75 & 51.53 & 5.8 \\

Aya-23-8B & 39.29 & 42.06 & 40.51 & 47.76 & 5.0 \\

\midrule
\textit{\textbf{European}} &  &  &  &  &  \\ \midrule

Mistral-7B & 57.05 & 56.89 & 48.82 & 53.08 & 2.5 \\

Occiglot-7B-eu5 & 45.93 & 47.26 & 34.61 & 51.09 & 3.8 \\

Salamandra-7B & 61.30 & 64.90 & 36.72 & 48.13 & 2.8 \\

\textbf{EuroLLM-9B} & \textbf{65.18} & \textbf{66.63} & \textbf{53.90} & \textbf{56.11} & \textbf{1.0} \\

\bottomrule
\end{tabular}%
}
\caption{ Comparison of the pre-trained versions of open-weight LLMs on Bulgarian benchmarks. }
\label{tab:pretrained_bg}
\end{table}


\begin{table}[H]
\resizebox{\textwidth}{!}{%
\begin{tabular}{lccccccc}
\toprule
Post-trained & \textbf{Arc} & \textbf{Hellaswag} & \textbf{MMLU} & \textbf{TruthfulQA} & \textbf{WMT24++} & \textbf{WMT24++} & \textbf{Borda C} $\downarrow$ \\
    &  &  &  &  & \footnotesize{en→xx} & \footnotesize{xx→en} &  \\
    & (25-shot) & (10-shot) & (5-shot) & (0-shot) & (0-shot) & (0-shot) &  \\ \midrule
\textit{\textbf{Non-European}}  &  &  &  &  &  &  \\ \midrule
Gemma-2-9B-IT & \textbf{64.51} & \textbf{63.42} & \textbf{64.96} & \textbf{61.71} & \textbf{80.74} & \textbf{82.99} & \textbf{1.0} \\
LLaMa-3.1-8B-IT & 51.47 & 54.24 & 51.27 & 54.54 & 75.99 & 81.19 & 2.7 \\
Granite-3-8B-IT & 43.11 & 46.51 & 42.11 & 57.79 & 61.20 & 79.87 & 4.0 \\
Qwen-2.5-7B-IT & 47.70 & 49.97 & 61.18 & 58.56 & 62.96 & 81.50 & 2.5 \\
OLMo-2-7B-IT & 34.96 & 38.84 & 37.25 & 52.24 & 53.98 & 76.78 & 6.0 \\
Aya-Expanse-8B & 42.26 & 45.23 & 48.43 & 52.99 & 60.44 & 79.29 & 4.8 \\

\midrule
\textit{\textbf{European}}  &  &  &  &  &  &  \\ \midrule
Mistral-7B-IT & 58.78 & 57.14 & 48.23 & \textbf{61.68} & 74.39 & 80.76 & 3.2 \\
Ministral-8B-IT & 52.44 & 53.35 & 48.78 & 54.17 & 69.91 & 79.96 & 4.7 \\
Occiglot-7B-eu5-IT & 39.94 & 46.11 & 35.82 & 53.14 & 57.66 & 70.93 & 6.7 \\
Salamandra-7B-IT & 55.12 & 63.56 & 46.23 & 53.35 & 83.19 & 83.56 & 3.0 \\
Pharia-1-LLM-7B-C & 25.72 & 31.02 & 24.76 & 50.80 & 34.45 & 57.66 & 8.0 \\
Teuken-7B-IT-R-v0.4 & 54.23 & 61.10 & 30.74 & 55.41 & 71.63 & 76.26 & 4.7 \\
Teuken-7B-IT-C-v0.4 & 51.46 & 59.68 & 36.80 & 53.03 & 74.41 & 81.27 & 4.7 \\
\textbf{EuroLLM-9B-IT} & \textbf{59.92} & \textbf{63.69} & \textbf{53.92} & 57.75 & \textbf{85.65} & \textbf{84.06} & \textbf{1.2} \\

\bottomrule
\end{tabular}%
}
\caption{ Comparison of the post-trained versions of open-weight LLMs on Bulgarian benchmarks. }
\label{tab:posttrained_bg}
\end{table}

\newpage

\subsubsection{Czech (cs)}

\begin{table}[H]
\resizebox{\textwidth}{!}{%
\begin{tabular}{lccccc}
\toprule
Pre-trained & \textbf{Arc} & \textbf{Hellaswag} & \textbf{MMLU} & \textbf{TruthfulQA} & \textbf{Borda C}  $\downarrow$ \\
 & (25-shot) & (10-shot) & (5-shot) & (0-shot) &  \\ \midrule
\textit{\textbf{Non-European}} &  &  &  &  &  \\ \midrule

Gemma-2-9B & \textbf{67.56} & \textbf{66.34} & \textbf{66.14} & 50.73 & \textbf{1.5} \\

LLaMa-3.1-8B & 56.81 & 58.31 & 55.51 & 50.70 & 3.5 \\

Granite-3-8B & 56.94 & 60.95 & 52.02 & 52.40 & 2.8 \\

Qwen-2.5-7B & 51.95 & 51.92 & 63.58 & \textbf{54.81} & 3.2 \\

OLMo-2-7B & 33.42 & 35.44 & 38.99 & 43.51 & 6.0 \\

Aya-23-8B & 56.58 & 61.10 & 49.93 & 46.70 & 4.0 \\

\midrule
\textit{\textbf{European}} &  &  &  &  &  \\ \midrule

Mistral-7B & 55.93 & 53.88 & 51.86 & 48.71 & 2.5 \\

Occiglot-7B-eu5 & 46.34 & 47.64 & 43.31 & 44.90 & 4.0 \\

Salamandra-7B & 61.41 & 63.83 & 43.59 & 47.39 & 2.5 \\

\textbf{EuroLLM-9B} & \textbf{66.51} & \textbf{65.88} & \textbf{55.82} & \textbf{51.08} & \textbf{1.0} \\

\bottomrule
\end{tabular}%
}
\caption{ Comparison of the pre-trained versions of open-weight LLMs on Czech benchmarks. }
\label{tab:pretrained_cs}
\end{table}


\begin{table}[H]
\resizebox{\textwidth}{!}{%
\begin{tabular}{lccccccc}
\toprule
Post-trained & \textbf{Arc} & \textbf{Hellaswag} & \textbf{MMLU} & \textbf{TruthfulQA} & \textbf{WMT24++} & \textbf{WMT24++} & \textbf{Borda C} $\downarrow$ \\
    &  &  &  &  & \footnotesize{en→xx} & \footnotesize{xx→en} &  \\
    & (25-shot) & (10-shot) & (5-shot) & (0-shot) & (0-shot) & (0-shot) &  \\ \midrule
\textit{\textbf{Non-European}}  &  &  &  &  &  &  \\ \midrule
Gemma-2-9B-IT & \textbf{63.98} & 62.87 & \textbf{65.76} & \textbf{61.10} & 81.27 & 82.80 & \textbf{1.7} \\
LLaMa-3.1-8B-IT & 52.48 & 56.23 & 58.12 & 54.70 & 77.68 & 80.90 & 4.2 \\
Granite-3-8B-IT & 56.67 & 62.46 & 54.76 & 58.76 & 79.33 & 82.84 & 2.8 \\
Qwen-2.5-7B-IT & 48.32 & 51.87 & 63.75 & 57.24 & 71.05 & 81.99 & 4.0 \\
OLMo-2-7B-IT & 33.31 & 36.16 & 39.48 & 42.75 & 49.59 & 75.48 & 6.0 \\
Aya-Expanse-8B & 55.63 & \textbf{62.89} & 55.39 & 54.94 & \textbf{84.23} & \textbf{83.43} & 2.3 \\

\midrule
\textit{\textbf{European}}  &  &  &  &  &  &  \\ \midrule
Mistral-7B-IT & 56.86 & 53.90 & 50.48 & \textbf{60.48} & 71.90 & 80.20 & 3.5 \\
Ministral-8B-IT & 50.68 & 51.76 & 52.15 & 53.07 & 70.29 & 80.31 & 4.8 \\
Occiglot-7B-eu5-IT & 40.43 & 46.31 & 40.59 & 47.55 & 59.05 & 71.08 & 6.7 \\
Salamandra-7B-IT & 54.82 & 62.11 & 47.76 & 51.50 & 81.64 & 83.33 & 3.2 \\
Pharia-1-LLM-7B-C & 27.62 & 30.22 & 29.29 & 45.25 & 35.10 & 59.20 & 8.0 \\
Teuken-7B-IT-R-v0.4 & 54.56 & 60.57 & 36.89 & 54.23 & 77.79 & 76.02 & 4.2 \\
Teuken-7B-IT-C-v0.4 & 52.55 & 58.91 & 38.06 & 53.14 & 75.90 & 81.38 & 4.3 \\
\textbf{EuroLLM-9B-IT} & \textbf{60.58} & \textbf{64.40} & \textbf{56.08} & 53.35 & \textbf{84.91} & \textbf{83.69} & \textbf{1.3} \\

\bottomrule
\end{tabular}%
}
\caption{ Comparison of the post-trained versions of open-weight LLMs on Czech benchmarks. }
\label{tab:posttrained_cs}
\end{table}

\newpage

\subsubsection{Danish (da)}

\begin{table}[H]
\resizebox{\textwidth}{!}{%
\begin{tabular}{lccccc}
\toprule
Pre-trained & \textbf{Arc} & \textbf{Hellaswag} & \textbf{MMLU} & \textbf{TruthfulQA} & \textbf{Borda C}  $\downarrow$ \\
 & (25-shot) & (10-shot) & (5-shot) & (0-shot) &  \\ \midrule
\textit{\textbf{Non-European}} &  &  &  &  &  \\ \midrule

Gemma-2-9B & \textbf{69.10} & \textbf{69.78} & \textbf{68.05} & 50.60 & \textbf{1.2} \\

LLaMa-3.1-8B & 55.90 & 61.49 & 56.92 & 50.53 & 2.5 \\

Granite-3-8B & 44.30 & 49.73 & 49.11 & 50.49 & 4.0 \\

Qwen-2.5-7B & 50.90 & 52.38 & 65.41 & \textbf{53.10} & 2.2 \\

OLMo-2-7B & 35.93 & 42.31 & 44.92 & 42.78 & 5.8 \\

Aya-23-8B & 39.40 & 47.67 & 43.89 & 44.45 & 5.2 \\

\midrule
\textit{\textbf{European}} &  &  &  &  &  \\ \midrule

Mistral-7B & 55.19 & 57.98 & 54.25 & 45.60 & 2.5 \\

Occiglot-7B-eu5 & 47.48 & 52.81 & 44.69 & 42.89 & 3.8 \\

Salamandra-7B & 61.32 & 66.75 & 43.37 & 43.97 & 2.8 \\

\textbf{EuroLLM-9B} & \textbf{66.17} & \textbf{68.69} & \textbf{56.74} & \textbf{50.89} & \textbf{1.0} \\

\bottomrule
\end{tabular}%
}
\caption{ Comparison of the pre-trained versions of open-weight LLMs on Danish benchmarks. }
\label{tab:pretrained_da}
\end{table}


\begin{table}[H]
\resizebox{\textwidth}{!}{%
\begin{tabular}{lccccccc}
\toprule
Post-trained & \textbf{Arc} & \textbf{Hellaswag} & \textbf{MMLU} & \textbf{TruthfulQA} & \textbf{WMT24++} & \textbf{WMT24++} & \textbf{Borda C} $\downarrow$ \\
    &  &  &  &  & \footnotesize{en→xx} & \footnotesize{xx→en} &  \\
    & (25-shot) & (10-shot) & (5-shot) & (0-shot) & (0-shot) & (0-shot) &  \\ \midrule
\textit{\textbf{Non-European}}  &  &  &  &  &  &  \\ \midrule
Gemma-2-9B-IT & \textbf{65.09} & \textbf{66.25} & \textbf{67.07} & \textbf{60.44} & \textbf{82.54} & \textbf{84.74} & \textbf{1.0} \\
LLaMa-3.1-8B-IT & 50.65 & 59.55 & 58.95 & 56.01 & 79.93 & 82.80 & 2.5 \\
Granite-3-8B-IT & 44.56 & 51.63 & 50.67 & 52.81 & 63.92 & 81.67 & 4.3 \\
Qwen-2.5-7B-IT & 48.06 & 52.17 & 64.92 & 57.16 & 68.41 & 83.24 & 2.5 \\
OLMo-2-7B-IT & 36.41 & 43.28 & 44.32 & 45.28 & 59.15 & 80.12 & 6.0 \\
Aya-Expanse-8B & 40.58 & 50.59 & 50.55 & 49.61 & 66.16 & 82.37 & 4.7 \\

\midrule
\textit{\textbf{European}}  &  &  &  &  &  &  \\ \midrule
Mistral-7B-IT & 57.25 & 58.52 & 52.13 & \textbf{58.26} & 75.24 & 82.50 & 3.8 \\
Ministral-8B-IT & 51.11 & 56.97 & 53.41 & 51.36 & 76.06 & 82.88 & 4.5 \\
Occiglot-7B-eu5-IT & 43.18 & 50.36 & 41.01 & 47.25 & 61.03 & 74.35 & 6.8 \\
Salamandra-7B-IT & 56.49 & 65.53 & 47.80 & 50.31 & 82.79 & 85.21 & 3.3 \\
Pharia-1-LLM-7B-C & 31.23 & 35.79 & 32.94 & 40.05 & 47.15 & 71.98 & 8.0 \\
Teuken-7B-IT-R-v0.4 & 57.28 & 63.62 & 40.49 & 52.24 & 76.83 & 79.73 & 4.2 \\
Teuken-7B-IT-C-v0.4 & 53.99 & 62.21 & 41.52 & 51.33 & 78.95 & 83.26 & 4.2 \\
\textbf{EuroLLM-9B-IT} & \textbf{60.81} & \textbf{66.80} & \textbf{55.61} & 55.05 & \textbf{85.14} & \textbf{85.61} & \textbf{1.2} \\

\bottomrule
\end{tabular}%
}
\caption{ Comparison of the post-trained versions of open-weight LLMs on Danish benchmarks. }
\label{tab:posttrained_da}
\end{table}

\newpage
\subsubsection{Dutch (nl)}

\begin{table}[H]
\resizebox{\textwidth}{!}{%
\begin{tabular}{lcccccccc}
\toprule
Pre-trained & \textbf{Arc} & \textbf{Hellaswag} & \textbf{MMLU} & \textbf{TruthfulQA} & \textbf{MMLU-pro} & \textbf{MUSR} & \textbf{Borda C}  $\downarrow$ \\
 & (25-shot) & (10-shot) & (5-shot) & (0-shot) & (5-shot) & (0-shot) &  \\ \midrule
\textit{\textbf{Non-European}} &  &  &  &  &  &  &  \\ \midrule

Gemma-2-9B & \textbf{69.74} & \textbf{68.86} & \textbf{67.63} & 53.24 & 29.01 & \textbf{10.25} & \textbf{1.5} \\

LLaMa-3.1-8B & 58.10 & 63.10 & 58.60 & 52.11 & 19.50 & 4.10 & 3.6 \\

Granite-3-8B & 60.74 & 66.95 & 56.30 & 54.95 & 19.04 & 5.39 & 2.8 \\

Qwen-2.5-7B & 56.51 & 58.41 & 67.62 & \textbf{57.67} & \textbf{29.97} & 5.69 & 2.7 \\

OLMo-2-7B & 38.39 & 43.31 & 45.41 & 45.35 & 11.97 & 3.67 & 6.0 \\

Aya-23-8B & 57.36 & 63.95 & 50.93 & 51.33 & 12.64 & 4.10 & 4.4 \\

\midrule
\textit{\textbf{European}} &  &  &  &  &  &  &  \\ \midrule

Mistral-7B & 57.02 & 58.34 & 54.61 & 51.23 & 16.38 & \textbf{12.63} & 2.2 \\

Occiglot-7B-eu5 & 51.01 & 54.93 & 46.56 & 46.56 & 10.38 & 2.81 & 3.5 \\

Salamandra-7B & 61.77 & 65.20 & 44.87 & 48.19 & 5.34 & 2.68 & 3.2 \\

\textbf{EuroLLM-9B} & \textbf{67.19} & \textbf{67.69} & \textbf{56.57} & \textbf{56.37} & \textbf{17.18} & 9.63 & \textbf{1.2} \\

\bottomrule
\end{tabular}%
}
\caption{ Comparison of the pre-trained versions of open-weight LLMs on Dutch benchmarks. }
\label{tab:pretrained_nl}
\end{table}


\begin{table}[H]
\resizebox{\textwidth}{!}{%
\begin{tabular}{lccccccccc}
\toprule
Post-trained & \textbf{Arc} & \textbf{Hellaswag} & \textbf{MMLU} & \textbf{TruthfulQA} & \textbf{MUSR} & \textbf{MMLU-pro} & \textbf{WMT24++} & \textbf{WMT24++} & \textbf{Borda C} $\downarrow$ \\
    &  &  &  &  &  &  & \footnotesize{en→xx} & \footnotesize{xx→en} &  \\
    & (25-shot) & (10-shot) & (5-shot) & (0-shot) & (0-shot) & (5-shot) & (0-shot) & (0-shot) &  \\ \midrule
\textit{\textbf{Non-European}}  &  &  &  &  &  &  &  &  \\ \midrule
Gemma-2-9B-IT & \textbf{65.65} & 66.92 & \textbf{67.52} & 60.47 & 27.68 & \textbf{9.89} & 81.79 & 84.17 & \textbf{1.9} \\
LLaMa-3.1-8B-IT & 52.79 & 61.91 & 60.67 & 56.20 & 22.66 & 3.57 & 80.65 & 82.90 & 4.1 \\
Granite-3-8B-IT & 59.18 & \textbf{68.89} & 58.31 & \textbf{62.89} & 18.61 & 9.26 & 81.21 & 84.18 & 2.4 \\
Qwen-2.5-7B-IT & 51.90 & 58.08 & 66.74 & 61.42 & \textbf{28.23} & 7.67 & 77.05 & 83.51 & 3.4 \\
OLMo-2-7B-IT & 39.97 & 44.90 & 45.08 & 48.22 & 9.93 & 3.08 & 66.60 & 81.00 & 6.0 \\
Aya-Expanse-8B & 56.14 & 65.80 & 56.36 & 58.25 & 19.54 & 6.65 & \textbf{83.02} & \textbf{84.42} & 2.8 \\

\midrule
\textit{\textbf{European}}  &  &  &  &  &  &  &  &  \\ \midrule
Mistral-7B-IT & 58.68 & 59.07 & 52.77 & \textbf{61.56} & \textbf{17.51} & 6.58 & 76.26 & 81.98 & 4.1 \\
Ministral-8B-IT & 52.85 & 56.50 & 53.77 & 49.54 & 16.34 & 4.23 & 76.80 & 82.66 & 4.9 \\
Occiglot-7B-eu5-IT & 43.16 & 52.83 & 44.31 & 49.58 & 10.31 & 2.94 & 65.82 & 73.52 & 6.9 \\
Salamandra-7B-IT & 55.26 & 64.52 & 48.44 & 53.42 & 5.76 & 7.27 & 81.31 & 84.27 & 3.9 \\
Pharia-1-LLM-7B-C & 55.02 & 64.04 & 44.33 & 39.35 & 10.06 & \textbf{9.09} & 77.99 & 82.40 & 4.6 \\
Teuken-7B-IT-R-v0.4 & 57.28 & 62.24 & 41.44 & 55.51 & 9.12 & 1.29 & 73.25 & 79.52 & 6.0 \\
Teuken-7B-IT-C-v0.4 & 55.65 & 62.19 & 42.38 & 56.45 & 9.22 & 3.44 & 79.57 & 82.67 & 4.6 \\
\textbf{EuroLLM-9B-IT} & \textbf{61.17} & \textbf{67.05} & \textbf{56.17} & 57.10 & 15.48 & 8.90 & \textbf{84.13} & \textbf{84.86} & \textbf{1.5} \\

\bottomrule
\end{tabular}%
}
\caption{ Comparison of the post-trained versions of open-weight LLMs on Dutch benchmarks. }
\label{tab:posttrained_nl}
\end{table}

\newpage

\subsubsection{English (en)}

\begin{table}[H]
\resizebox{\textwidth}{!}{%
\begin{tabular}{lcccccccc}
\toprule
Pre-trained & \textbf{Arc} & \textbf{Hellaswag} & \textbf{MMLU} & \textbf{TruthfulQA} & \textbf{MMLU-pro} & \textbf{MUSR} & \textbf{Borda C}  $\downarrow$ \\
 & (25-shot) & (10-shot) & (5-shot) & (0-shot) & (5-shot) & (0-shot) &  \\ \midrule
\textit{\textbf{Non-European}} &  &  &  &  &  &  &  \\ \midrule

Gemma-2-9B & \textbf{78.66} & 82.39 & 70.30 & 45.44 & 34.87 & \textbf{14.48} & \textbf{1.8} \\

LLaMa-3.1-8B & 71.12 & 82.08 & 65.10 & 45.21 & 25.17 & 9.13 & 4.3 \\

Granite-3-8B & 75.45 & \textbf{83.51} & 63.40 & 51.31 & 25.82 & 9.36 & 2.8 \\

Qwen-2.5-7B & 75.47 & 80.01 & \textbf{74.26} & \textbf{56.31} & \textbf{37.33} & 13.06 & 2.0 \\

OLMo-2-7B & 75.27 & 82.23 & 62.95 & 43.31 & 22.74 & 10.12 & 4.3 \\

Aya-23-8B & 66.84 & 77.83 & 54.89 & 45.35 & 16.68 & 5.85 & 5.7 \\

\midrule
\textit{\textbf{European}} &  &  &  &  &  &  &  \\ \midrule

Mistral-7B & 72.76 & \textbf{83.03} & \textbf{62.33} & 42.59 & \textbf{21.78} & 8.50 & \textbf{1.5} \\

Occiglot-7B-eu5 & 67.38 & 78.96 & 52.81 & 40.32 & 13.87 & 2.68 & 3.2 \\

Salamandra-7B & 70.92 & 77.85 & 46.48 & 41.49 & 5.52 & 2.58 & 3.7 \\

\textbf{EuroLLM-9B} & \textbf{72.88} & 78.63 & 57.41 & \textbf{48.50} & 17.68 & \textbf{12.47} & 1.7 \\

\bottomrule
\end{tabular}%
}
\caption{ Comparison of the pre-trained versions of open-weight LLMs on English benchmarks. }
\label{tab:pretrained_en}
\end{table}


\begin{table}[H]
\resizebox{\textwidth}{!}{%
\begin{tabular}{lcccccccc}
\toprule
Post-trained & \textbf{Arc} & \textbf{Hellaswag} & \textbf{MMLU} & \textbf{TruthfulQA} & \textbf{MMLU-pro} & \textbf{MUSR} & \textbf{IFEval} & \textbf{Borda C}  $\downarrow$ \\
 & (25-shot) & (10-shot) & (5-shot) & (0-shot) & (5-shot) & (0-shot) & (0-shot) &  \\ \midrule
\textit{\textbf{Non-European}} &  &  &  &  &  &  &  &   \\ \midrule

Gemma-2-9B-IT & \textbf{78.23} & 78.87 & 71.87 & 60.49 & 31.77 & \textbf{9.16} & 75.00 & \textbf{2.3} \\

LLaMa-3.1-8B-IT & 66.32 & 77.39 & 68.25 & 52.99 & 30.52 & 7.21 & \textbf{77.83} & 3.6 \\

Granite-3-8B-IT & 71.72 & \textbf{84.40} & 65.46 & \textbf{66.35} & 27.30 & 4.63 & 54.38 & 3.4 \\

Qwen-2.5-7B-IT & 67.96 & 73.46 & \textbf{73.41} & 62.45 & \textbf{36.91} & 9.00 & 76.16 & 2.6 \\

OLMo-2-7B-IT & 70.13 & 80.32 & 61.00 & 55.58 & 20.17 & 6.08 & 76.07 & 4.0 \\

Aya-Expanse-8B & 66.11 & 78.37 & 62.01 & 52.28 & 22.27 & 4.86 & 63.95 & 4.6 \\

\midrule
\textit{\textbf{European}} &  &  &  &  &  &  &  &  \\ \midrule

Mistral-7B-IT & \textbf{75.79} & \textbf{82.00} & 60.67 & \textbf{66.01} & \textbf{23.21} & 3.54 & 53.43 & 2.9 \\

Ministral-8B-IT & 68.86 & 78.53 & \textbf{62.93} & 54.02 & 13.12 & 7.84 & 58.57 & \textbf{2.4} \\

Occiglot-7B-eu5-IT & 62.07 & 76.58 & 50.65 & 47.95 & 13.89 & 7.21 & 40.09 & 5.0 \\

Salamandra-7B-IT & 66.91 & 76.40 & 52.54 & 50.87 & 9.08 & 8.17 & 24.09 & 5.6 \\

Pharia-1-LLM-7B-C & 68.41 & 75.15 & 49.32 & 45.88 & 11.86 & \textbf{9.75} & 36.95 & 5.0 \\

Teuken-7B-IT-R-v0.4 & 67.47 & 73.32 & 45.53 & 51.36 & 10.82 & 4.07 & 35.26 & 6.0 \\

Teuken-7B-IT-C-v0.4 & 67.61 & 73.51 & 45.33 & 50.93 & 10.78 & 1.72 & 35.05 & 6.6 \\

\textbf{EuroLLM-9B-IT} & 68.68 & 76.38 & 58.36 & 49.77 & 19.58 & 8.44 & \textbf{65.69} & 3.1 \\

\bottomrule
\end{tabular}%
}
\caption{ Comparison of the post-trained versions of open-weight LLMs on English benchmarks. }
\label{tab:posttrained_en}
\end{table}

\newpage

\subsubsection{Estonian (et)}

\begin{table}[H]
\resizebox{\textwidth}{!}{%
\begin{tabular}{lccccc}
\toprule
Pre-trained & \textbf{Arc} & \textbf{Hellaswag} & \textbf{MMLU} & \textbf{TruthfulQA} & \textbf{Borda C}  $\downarrow$ \\
 & (25-shot) & (10-shot) & (5-shot) & (0-shot) &  \\ \midrule
\textit{\textbf{Non-European}} &  &  &  &  &  \\ \midrule

Gemma-2-9B & \textbf{59.88} & \textbf{59.22} & \textbf{62.50} & \textbf{49.50} & \textbf{1.0} \\

LLaMa-3.1-8B & 44.35 & 46.33 & 48.58 & 46.38 & 2.5 \\

Granite-3-8B & 28.91 & 32.24 & 35.24 & 44.11 & 4.2 \\

Qwen-2.5-7B & 34.29 & 37.08 & 51.68 & 48.89 & 2.5 \\

OLMo-2-7B & 26.78 & 31.14 & 32.02 & 43.40 & 6.0 \\

Aya-23-8B & 28.07 & 32.76 & 33.38 & 43.43 & 4.8 \\

\midrule
\textit{\textbf{European}} &  &  &  &  &  \\ \midrule

Mistral-7B & 29.64 & 34.15 & 38.85 & 43.15 & 3.2 \\

Occiglot-7B-eu5 & 29.07 & 33.37 & 32.97 & 44.19 & 3.8 \\

Salamandra-7B & 56.08 & 58.92 & 41.66 & 45.42 & 2.0 \\

\textbf{EuroLLM-9}B & \textbf{62.37} & \textbf{62.28} & \textbf{54.59} & \textbf{47.15} & \textbf{1.0} \\

\bottomrule
\end{tabular}%
}
\caption{ Comparison of the pre-trained versions of open-weight LLMs on Estonian benchmarks. }
\label{tab:pretrained_et}
\end{table}


\begin{table}[H]
\resizebox{\textwidth}{!}{%
\begin{tabular}{lccccccc}
\toprule
Post-trained & \textbf{Arc} & \textbf{Hellaswag} & \textbf{MMLU} & \textbf{TruthfulQA} & \textbf{WMT24++} & \textbf{WMT24++} & \textbf{Borda C} $\downarrow$ \\
    &  &  &  &  & \footnotesize{en→xx} & \footnotesize{xx→en} &  \\
    & (25-shot) & (10-shot) & (5-shot) & (0-shot) & (0-shot) & (0-shot) &  \\ \midrule
\textit{\textbf{Non-European}}  &  &  &  &  &  &  \\ \midrule
Gemma-2-9B-IT & \textbf{55.71} & \textbf{55.04} & \textbf{61.44} & \textbf{57.02} & \textbf{79.42} & \textbf{83.75} & \textbf{1.0} \\
LLaMa-3.1-8B-IT & 41.21 & 45.35 & 51.10 & 50.13 & 72.95 & 79.33 & 2.2 \\
Granite-3-8B-IT & 29.11 & 31.89 & 36.60 & 46.62 & 42.63 & 67.92 & 5.0 \\
Qwen-2.5-7B-IT & 33.05 & 37.40 & 51.75 & 49.33 & 53.59 & 76.27 & 3.0 \\
OLMo-2-7B-IT & 27.84 & 30.87 & 32.61 & 41.92 & 43.09 & 64.51 & 5.8 \\
Aya-Expanse-8B & 28.04 & 32.04 & 38.83 & 49.94 & 45.59 & 70.87 & 4.0 \\

\midrule
\textit{\textbf{European}}  &  &  &  &  &  &  \\ \midrule
Mistral-7B-IT & 30.82 & 33.72 & 36.12 & 46.72 & 47.27 & 70.13 & 6.0 \\
Ministral-8B-IT & 40.43 & 43.81 & 42.45 & \textbf{51.58} & 66.52 & 78.32 & 4.0 \\
Occiglot-7B-eu5-IT & 28.18 & 32.51 & 28.44 & 46.53 & 44.04 & 63.49 & 7.0 \\
Salamandra-7B-IT & 49.21 & 57.87 & 45.53 & 49.22 & 83.26 & 84.94 & 2.5 \\
Pharia-1-LLM-7B-C & 25.53 & 29.85 & 28.32 & 44.77 & 36.46 & 51.55 & 8.0 \\
Teuken-7B-IT-R-v0.4 & 47.15 & 53.39 & 37.94 & 54.29 & 77.81 & 71.30 & 3.3 \\
Teuken-7B-IT-C-v0.4 & 44.68 & 52.45 & 39.72 & 50.53 & 72.39 & 81.90 & 3.8 \\
\textbf{EuroLLM-9B-IT} & \textbf{55.48} & \textbf{58.96} & \textbf{54.09} & 51.55 & \textbf{86.49} & \textbf{85.81} & \textbf{1.3} \\

\bottomrule
\end{tabular}%
}
\caption{ Comparison of the post-trained versions of open-weight LLMs on Estonian benchmarks. }
\label{tab:posttrained_et}
\end{table}

\newpage
\subsubsection{German (de)}

\begin{table}[H]
\resizebox{\textwidth}{!}{%
\begin{tabular}{lcccccccc}
\toprule
Pre-trained & \textbf{Arc} & \textbf{Hellaswag} & \textbf{MMLU} & \textbf{TruthfulQA} & \textbf{MMLU-pro} & \textbf{MUSR} & \textbf{Borda C}  $\downarrow$ \\
 & (25-shot) & (10-shot) & (5-shot) & (0-shot) & (5-shot) & (0-shot) &  \\ \midrule
\textit{\textbf{Non-European}} &  &  &  &  &  &  &  \\ \midrule

Gemma-2-9B & \textbf{69.71} & \textbf{69.15} & 67.83 & 48.71 & 28.98 & \textbf{10.65} & \textbf{2.0} \\

LLaMa-3.1-8B & 58.69 & 63.36 & 59.09 & 49.57 & 19.57 & 8.04 & 3.8 \\

Granite-3-8B & 63.10 & 66.77 & 57.21 & 50.90 & 20.58 & 5.99 & 3.0 \\

Qwen-2.5-7B & 60.65 & 59.78 & \textbf{68.15} & \textbf{54.86} & \textbf{30.27} & 10.25 & 2.2 \\

OLMo-2-7B & 47.16 & 48.13 & 47.81 & 42.56 & 12.76 & 7.21 & 5.7 \\

Aya-23-8B & 59.99 & 64.48 & 51.88 & 50.08 & 14.26 & 4.10 & 4.3 \\

\midrule
\textit{\textbf{European}} &  &  &  &  &  &  &  \\ \midrule

Mistral-7B & 59.88 & 59.90 & 55.24 & 46.44 & 17.30 & 10.48 & 2.7 \\

Occiglot-7B-eu5 & 61.44 & \textbf{67.87} & 51.13 & 43.94 & 12.70 & 4.99 & 2.8 \\

Salamandra-7B & 62.18 & 65.73 & 43.82 & 44.11 & 4.50 & 3.97 & 3.3 \\

\textbf{EuroLLM-9B} & \textbf{66.74} & 67.28 & \textbf{56.53} & \textbf{51.80} & \textbf{17.62} & \textbf{12.17} & \textbf{1.2} \\

\bottomrule
\end{tabular}%
}
\caption{ Comparison of the pre-trained versions of open-weight LLMs on German benchmarks. }
\label{tab:pretrained_de}
\end{table}


\begin{table}[H]
\resizebox{\textwidth}{!}{%
\begin{tabular}{lccccccccc}
\toprule
Post-trained & \textbf{Arc} & \textbf{Hellaswag} & \textbf{MMLU} & \textbf{TruthfulQA} & \textbf{MUSR} & \textbf{MMLU-pro} & \textbf{WMT24++} & \textbf{WMT24++} & \textbf{Borda C} $\downarrow$ \\
    &  &  &  &  &  &  & \footnotesize{en→xx} & \footnotesize{xx→en} &  \\
    & (25-shot) & (10-shot) & (5-shot) & (0-shot) & (0-shot) & (5-shot) & (0-shot) & (0-shot) &  \\ \midrule
\textit{\textbf{Non-European}}  &  &  &  &  &  &  &  &  \\ \midrule
Gemma-2-9B-IT & \textbf{67.42} & 66.33 & 67.08 & 60.01 & 27.91 & \textbf{10.68} & 78.10 & 83.94 & \textbf{2.1} \\
LLaMa-3.1-8B-IT & 53.38 & 60.99 & 61.18 & 56.70 & 23.14 & 4.57 & 78.38 & 82.63 & 4.4 \\
Granite-3-8B-IT & 61.81 & \textbf{68.77} & 59.23 & \textbf{62.06} & 18.64 & 10.02 & 79.54 & 83.66 & 2.5 \\
Qwen-2.5-7B-IT & 57.52 & 58.83 & \textbf{67.27} & 60.76 & \textbf{29.17} & 10.32 & 76.27 & 83.52 & 2.9 \\
OLMo-2-7B-IT & 47.31 & 51.31 & 47.37 & 50.43 & 11.36 & 5.09 & 73.66 & 82.74 & 5.8 \\
Aya-Expanse-8B & 57.47 & 66.28 & 56.93 & 57.60 & 19.03 & 8.47 & \textbf{81.61} & \textbf{84.14} & 2.9 \\

\midrule
\textit{\textbf{European}}  &  &  &  &  &  &  &  &  \\ \midrule
Mistral-7B-IT & 61.03 & 60.43 & 53.11 & \textbf{61.77} & \textbf{17.76} & 8.99 & 74.58 & 81.86 & 4.6 \\
Ministral-8B-IT & 58.08 & 63.56 & \textbf{57.61} & 50.77 & 17.53 & 9.10 & 78.88 & 83.47 & 3.6 \\
Occiglot-7B-eu5-IT & 54.96 & 65.85 & 49.09 & 47.64 & 12.38 & 3.94 & 70.59 & 74.87 & 5.9 \\
Salamandra-7B-IT & 56.74 & 65.01 & 48.83 & 49.27 & 5.81 & 9.69 & 79.58 & 83.89 & 4.5 \\
Pharia-1-LLM-7B-C & 60.56 & 65.50 & 45.06 & 46.38 & 10.41 & 9.42 & 78.33 & 82.22 & 4.8 \\
Teuken-7B-IT-R-v0.4 & \textbf{61.51} & 63.43 & 41.00 & 53.31 & 9.04 & 2.61 & 77.31 & 79.21 & 5.5 \\
Teuken-7B-IT-C-v0.4 & 59.52 & 62.34 & 42.63 & 52.23 & 9.70 & 1.95 & 77.93 & 82.32 & 5.6 \\
\textbf{EuroLLM-9B-IT} & 60.92 & \textbf{66.27} & 56.37 & 52.02 & 17.30 & \textbf{11.97} & \textbf{82.25} & \textbf{84.33} & \textbf{2.0} \\

\bottomrule
\end{tabular}%
}
\caption{ Comparison of the post-trained versions of open-weight LLMs on German benchmarks. }
\label{tab:posttrained_de}
\end{table}

\newpage

\subsubsection{Greek (el)}

\begin{table}[H]
\resizebox{\textwidth}{!}{%
\begin{tabular}{lccccc}
\toprule
Pre-trained & \textbf{Arc} & \textbf{Hellaswag} & \textbf{MMLU} & \textbf{TruthfulQA} & \textbf{Borda C}  $\downarrow$ \\
 & (25-shot) & (10-shot) & (5-shot) & (0-shot) &  \\ \midrule
\textit{\textbf{Non-European}} &  &  &  &  &  \\ \midrule

Gemma-2-9B & \textbf{64.67} & \textbf{65.87} & \textbf{63.71} & \textbf{53.06} & \textbf{1.0} \\

LLaMa-3.1-8B & 50.45 & 55.16 & 51.29 & 50.40 & 3.0 \\

Granite-3-8B & 32.50 & 39.72 & 34.39 & 45.96 & 5.2 \\

Qwen-2.5-7B & 39.17 & 41.54 & 51.21 & 50.50 & 3.5 \\

OLMo-2-7B & 27.09 & 33.94 & 29.68 & 47.46 & 5.8 \\

Aya-23-8B & 55.94 & 62.21 & 47.91 & 50.63 & 2.5 \\

\midrule
\textit{\textbf{European}} &  &  &  &  &  \\ \midrule

Mistral-7B & 31.69 & 38.32 & 36.20 & 43.94 & 3.2 \\

Occiglot-7B-eu5 & 28.83 & 35.47 & 31.42 & 44.01 & 3.8 \\

Salamandra-7B & 58.80 & 64.82 & 39.57 & 48.54 & 2.0 \\

\textbf{EuroLLM-9B} & \textbf{64.80} & \textbf{64.96} & \textbf{52.26} & \textbf{52.20} & \textbf{1.0} \\

\bottomrule
\end{tabular}%
}
\caption{ Comparison of the pre-trained versions of open-weight LLMs on Greek benchmarks. }
\label{tab:pretrained_el}
\end{table}


\begin{table}[H]
\resizebox{\textwidth}{!}{%
\begin{tabular}{lccccccc}
\toprule
Post-trained & \textbf{Arc} & \textbf{Hellaswag} & \textbf{MMLU} & \textbf{TruthfulQA} & \textbf{WMT24++} & \textbf{WMT24++} & \textbf{Borda C} $\downarrow$ \\
    &  &  &  &  & \footnotesize{en→xx} & \footnotesize{xx→en} &  \\
    & (25-shot) & (10-shot) & (5-shot) & (0-shot) & (0-shot) & (0-shot) &  \\ \midrule
\textit{\textbf{Non-European}}  &  &  &  &  &  &  \\ \midrule
Gemma-2-9B-IT & \textbf{60.48} & 62.60 & \textbf{62.24} & \textbf{61.46} & 80.95 & 82.35 & \textbf{1.5} \\
LLaMa-3.1-8B-IT & 48.41 & 53.54 & 44.05 & 57.00 & 77.33 & 80.73 & 3.2 \\
Granite-3-8B-IT & 33.35 & 40.22 & 35.51 & 50.69 & 57.75 & 78.16 & 5.0 \\
Qwen-2.5-7B-IT & 38.27 & 42.19 & 52.77 & 56.85 & 61.07 & 79.60 & 3.8 \\
OLMo-2-7B-IT & 30.49 & 34.07 & 32.75 & 48.64 & 48.93 & 74.71 & 6.0 \\
Aya-Expanse-8B & 54.55 & \textbf{64.09} & 54.35 & 57.82 & \textbf{85.27} & \textbf{84.76} & \textbf{1.5} \\

\midrule
\textit{\textbf{European}}  &  &  &  &  &  &  \\ \midrule
Mistral-7B-IT & 31.76 & 37.76 & 37.03 & 49.80 & 54.12 & 74.68 & 5.5 \\
Ministral-8B-IT & 46.89 & 49.89 & 40.14 & 51.24 & 71.82 & 80.53 & 4.3 \\
Occiglot-7B-eu5-IT & 26.80 & 34.12 & 31.15 & 44.43 & 46.98 & 63.49 & 7.2 \\
Salamandra-7B-IT & 52.92 & 63.27 & 45.07 & 52.14 & 83.12 & \textbf{84.52} & 2.3 \\
Pharia-1-LLM-7B-C & 25.76 & 29.09 & 24.15 & 47.17 & 37.38 & 59.40 & 7.8 \\
Teuken-7B-IT-R-v0.4 & 54.19 & 60.91 & 34.22 & \textbf{57.41} & 77.32 & 73.55 & 3.3 \\
Teuken-7B-IT-C-v0.4 & 52.05 & 59.15 & 32.81 & 52.93 & 74.49 & 79.71 & 4.2 \\
\textbf{EuroLLM-9B-IT} & \textbf{58.23} & \textbf{63.34} & \textbf{51.67} & 54.64 & \textbf{85.88} & 84.20 & \textbf{1.3} \\

\bottomrule
\end{tabular}%
}
\caption{ Comparison of the post-trained versions of open-weight LLMs on Greek benchmarks. }
\label{tab:posttrained_el}
\end{table}

\newpage

\subsubsection{Finnish (fi)}

\begin{table}[H]
\resizebox{\textwidth}{!}{%
\begin{tabular}{lccccc}
\toprule
Pre-trained & \textbf{Arc} & \textbf{Hellaswag} & \textbf{MMLU} & \textbf{TruthfulQA} & \textbf{Borda C}  $\downarrow$ \\
 & (25-shot) & (10-shot) & (5-shot) & (0-shot) &  \\ \midrule
\textit{\textbf{Non-European}} &  &  &  &  &  \\ \midrule

Gemma-2-9B & \textbf{63.16} & \textbf{64.08} & \textbf{63.79} & 47.78 & \textbf{1.2} \\

LLaMa-3.1-8B & 48.55 & 52.29 & 51.32 & 45.33 & 2.5 \\

Granite-3-8B & 29.59 & 35.85 & 39.46 & 40.96 & 5.2 \\

Qwen-2.5-7B & 38.79 & 40.86 & 54.51 & \textbf{49.69} & 2.2 \\

OLMo-2-7B & 32.51 & 37.76 & 39.12 & 43.20 & 4.2 \\

Aya-23-8B & 29.92 & 34.95 & 34.96 & 42.86 & 5.5 \\

\midrule
\textit{\textbf{European}} &  &  &  &  &  \\ \midrule

Mistral-7B & 35.25 & 40.56 & 42.81 & 42.37 & 3.0 \\

Occiglot-7B-eu5 & 31.37 & 36.98 & 35.37 & 43.08 & 3.8 \\

Salamandra-7B & 56.10 & 61.39 & 41.72 & 43.41 & 2.2 \\

\textbf{EuroLLM-9B} & \textbf{63.93} & \textbf{65.46} & \textbf{54.79} & \textbf{48.88} & \textbf{1.0} \\

\bottomrule
\end{tabular}%
}
\caption{ Comparison of the pre-trained versions of open-weight LLMs on Finnish benchmarks. }
\label{tab:pretrained_fi}
\end{table}


\begin{table}[H]
\resizebox{\textwidth}{!}{%
\begin{tabular}{lccccccc}
\toprule
Post-trained & \textbf{Arc} & \textbf{Hellaswag} & \textbf{MMLU} & \textbf{TruthfulQA} & \textbf{WMT24++} & \textbf{WMT24++} & \textbf{Borda C} $\downarrow$ \\
    &  &  &  &  & \footnotesize{en→xx} & \footnotesize{xx→en} &  \\
    & (25-shot) & (10-shot) & (5-shot) & (0-shot) & (0-shot) & (0-shot) &  \\ \midrule
\textit{\textbf{Non-European}}  &  &  &  &  &  &  \\ \midrule
Gemma-2-9B-IT & \textbf{60.16} & \textbf{60.35} & \textbf{63.59} & \textbf{56.81} & \textbf{83.41} & \textbf{85.27} & \textbf{1.0} \\
LLaMa-3.1-8B-IT & 45.43 & 51.26 & 53.14 & 49.99 & 79.86 & 82.69 & 2.3 \\
Granite-3-8B-IT & 30.16 & 35.86 & 39.61 & 46.98 & 52.79 & 76.61 & 5.5 \\
Qwen-2.5-7B-IT & 37.22 & 40.98 & 54.82 & 51.10 & 63.73 & 80.76 & 2.8 \\
OLMo-2-7B-IT & 33.74 & 38.41 & 39.15 & 47.76 & 69.62 & 79.86 & 4.2 \\
Aya-Expanse-8B & 31.84 & 35.09 & 41.22 & 41.70 & 54.27 & 77.54 & 5.2 \\

\midrule
\textit{\textbf{European}}  &  &  &  &  &  &  \\ \midrule
Mistral-7B-IT & 35.79 & 40.11 & 40.68 & 47.79 & 59.82 & 78.86 & 5.7 \\
Ministral-8B-IT & 44.57 & 47.04 & 45.78 & 47.86 & 72.52 & 81.36 & 4.3 \\
Occiglot-7B-eu5-IT & 29.31 & 35.26 & 31.31 & 43.20 & 53.00 & 69.61 & 7.0 \\
Salamandra-7B-IT & 49.80 & 59.82 & 45.19 & 47.92 & 85.54 & 85.87 & 2.8 \\
Pharia-1-LLM-7B-C & 27.64 & 30.16 & 28.88 & 42.81 & 37.96 & 52.34 & 8.0 \\
Teuken-7B-IT-R-v0.4 & 51.14 & 57.60 & 38.66 & 50.80 & 78.41 & 78.34 & 3.7 \\
Teuken-7B-IT-C-v0.4 & 49.92 & 56.83 & 41.20 & 50.36 & 78.33 & 82.62 & 3.5 \\
\textbf{EuroLLM-9B-IT} & \textbf{58.34} & \textbf{61.70} & \textbf{52.91} & \textbf{52.57} & \textbf{87.74} & \textbf{86.23} & \textbf{1.0} \\

\bottomrule
\end{tabular}%
}
\caption{ Comparison of the post-trained versions of open-weight LLMs on Finnish benchmarks. }
\label{tab:posttrained_fi}
\end{table}

\newpage

\subsubsection{French (fr)}

\begin{table}[H]
\resizebox{\textwidth}{!}{%
\begin{tabular}{lcccccccc}
\toprule
Pre-trained & \textbf{Arc} & \textbf{Hellaswag} & \textbf{MMLU} & \textbf{TruthfulQA} & \textbf{MMLU-pro} & \textbf{MUSR} & \textbf{Borda C}  $\downarrow$ \\
 & (25-shot) & (10-shot) & (5-shot) & (0-shot) & (5-shot) & (0-shot) &  \\ \midrule
\textit{\textbf{Non-European}} &  &  &  &  &  &  &  \\ \midrule

Gemma-2-9B & \textbf{72.64} & \textbf{73.08} & 68.93 & 42.72 & 29.41 & 10.15 & 2.2 \\

LLaMa-3.1-8B & 62.20 & 67.45 & 59.67 & 47.79 & 19.23 & 2.94 & 4.0 \\

Granite-3-8B & 65.71 & 70.90 & 57.30 & 45.49 & 19.59 & 6.94 & 2.8 \\

Qwen-2.5-7B & 65.01 & 67.46 & \textbf{69.34} & \textbf{55.97} & \textbf{32.44} & \textbf{10.45} & \textbf{1.8} \\

OLMo-2-7B & 53.51 & 56.57 & 50.69 & 40.45 & 14.66 & 5.42 & 5.5 \\

Aya-23-8B & 61.40 & 68.78 & 51.97 & 44.15 & 13.82 & 3.47 & 4.7 \\

\midrule
\textit{\textbf{European}} &  &  &  &  &  &  &  \\ \midrule

Mistral-7B & 63.00 & 65.34 & 56.30 & 45.18 & 17.32 & 7.97 & 2.7 \\

Occiglot-7B-eu5 & 64.67 & \textbf{71.39} & 50.36 & 39.35 & 12.70 & 5.19 & 2.7 \\

Salamandra-7B & 64.63 & 68.96 & 44.29 & 40.33 & 5.38 & 3.27 & 3.5 \\

\textbf{EuroLLM-9B} & \textbf{68.95} & 70.18 & \textbf{56.44} & \textbf{46.56} & \textbf{18.36} & \textbf{11.08} & \textbf{1.2} \\

\bottomrule
\end{tabular}%
}
\caption{ Comparison of the pre-trained versions of open-weight LLMs on French benchmarks. }
\label{tab:pretrained_fr}
\end{table}


\begin{table}[H]
\resizebox{\textwidth}{!}{%
\begin{tabular}{lccccccccc}
\toprule
Post-trained & \textbf{Arc} & \textbf{Hellaswag} & \textbf{MMLU} & \textbf{TruthfulQA} & \textbf{MUSR} & \textbf{MMLU-pro} & \textbf{WMT24++} & \textbf{WMT24++} & \textbf{Borda C} $\downarrow$ \\
    &  &  &  &  &  &  & \footnotesize{en→xx} & \footnotesize{xx→en} &  \\
    & (25-shot) & (10-shot) & (5-shot) & (0-shot) & (0-shot) & (5-shot) & (0-shot) & (0-shot) &  \\ \midrule
\textit{\textbf{Non-European}}  &  &  &  &  &  &  &  &  \\ \midrule
Gemma-2-9B-IT & \textbf{68.71} & 69.91 & 67.81 & 60.81 & 28.76 & 9.66 & 77.99 & 83.21 & \textbf{2.4} \\
LLaMa-3.1-8B-IT & 56.94 & 65.01 & 61.95 & 56.55 & 23.92 & 4.30 & 78.00 & 82.35 & 4.4 \\
Granite-3-8B-IT & 63.64 & \textbf{72.01} & 60.26 & 63.95 & 20.03 & 11.28 & 78.88 & 83.21 & \textbf{2.4} \\
Qwen-2.5-7B-IT & 58.91 & 65.60 & \textbf{68.64} & \textbf{64.45} & \textbf{30.74} & \textbf{11.71} & 77.02 & 83.11 & 2.6 \\
OLMo-2-7B-IT & 53.14 & 59.07 & 50.56 & 50.81 & 13.20 & 3.64 & 76.69 & 82.67 & 5.9 \\
Aya-Expanse-8B & 60.10 & 69.59 & 57.51 & 57.57 & 20.24 & 5.36 & \textbf{80.83} & \textbf{83.72} & 2.8 \\

\midrule
\textit{\textbf{European}}  &  &  &  &  &  &  &  &  \\ \midrule
Mistral-7B-IT & \textbf{63.96} & 65.90 & 54.35 & \textbf{62.00} & 18.32 & 9.16 & 75.62 & 81.62 & 3.9 \\
Ministral-8B-IT & 59.92 & 67.75 & \textbf{57.08} & 55.17 & \textbf{18.82} & 8.01 & 78.53 & 82.83 & 3.2 \\
Occiglot-7B-eu5-IT & 58.18 & 69.00 & 48.86 & 46.99 & 12.04 & 4.80 & 70.77 & 74.87 & 6.0 \\
Salamandra-7B-IT & 58.64 & 67.98 & 48.69 & 47.92 & 7.07 & 5.92 & 79.14 & 83.57 & 4.9 \\
Pharia-1-LLM-7B-C & 61.65 & 68.36 & 44.99 & 47.13 & 10.41 & \textbf{11.38} & 77.04 & 81.82 & 4.2 \\
Teuken-7B-IT-R-v0.4 & 60.06 & 65.94 & 41.47 & 56.16 & 9.73 & 2.78 & 76.19 & 77.36 & 5.9 \\
Teuken-7B-IT-C-v0.4 & 59.42 & 65.72 & 43.57 & 53.63 & 10.93 & 1.46 & 76.97 & 80.51 & 6.1 \\
\textbf{EuroLLM-9B-IT} & 62.47 & \textbf{69.31} & 57.03 & 50.90 & 18.41 & 8.04 & \textbf{81.18} & \textbf{84.20} & \textbf{2.2} \\

\bottomrule
\end{tabular}%
}
\caption{ Comparison of the post-trained versions of open-weight LLMs on French benchmarks. }
\label{tab:posttrained_fr}
\end{table}

\newpage

\subsubsection{Hungarian (hu)}

\begin{table}[H]
\resizebox{\textwidth}{!}{%
\begin{tabular}{lccccc}
\toprule
Pre-trained & \textbf{Arc} & \textbf{Hellaswag} & \textbf{MMLU} & \textbf{TruthfulQA} & \textbf{Borda C}  $\downarrow$ \\
 & (25-shot) & (10-shot) & (5-shot) & (0-shot) &  \\ \midrule
\textit{\textbf{Non-European}} &  &  &  &  &  \\ \midrule

Gemma-2-9B & \textbf{63.07} & \textbf{63.55} & \textbf{64.10} & \textbf{53.46} & \textbf{1.0} \\

LLaMa-3.1-8B & 51.98 & 55.83 & 54.76 & 50.47 & 2.2 \\

Granite-3-8B & 29.96 & 33.48 & 37.84 & 50.17 & 4.2 \\

Qwen-2.5-7B & 36.23 & 38.99 & 53.59 & 51.64 & 2.8 \\

OLMo-2-7B & 28.44 & 31.78 & 33.81 & 47.96 & 6.0 \\

Aya-23-8B & 29.88 & 34.68 & 37.16 & 49.23 & 4.8 \\

\midrule
\textit{\textbf{European}} &  &  &  &  &  \\ \midrule

Mistral-7B & 49.42 & 49.50 & 50.07 & 47.92 & 2.8 \\

Occiglot-7B-eu5 & 40.83 & 41.74 & 40.50 & 47.02 & 4.0 \\

Salamandra-7B & 56.43 & 60.15 & 40.77 & 49.52 & 2.2 \\

\textbf{EuroLLM-9B} & \textbf{61.86} & \textbf{62.44} & \textbf{54.30} & \textbf{55.69} & \textbf{1.0} \\

\bottomrule
\end{tabular}%
}
\caption{ Comparison of the pre-trained versions of open-weight LLMs on Hungarian benchmarks. }
\label{tab:pretrained_hu}
\end{table}


\begin{table}[H]
\resizebox{\textwidth}{!}{%
\begin{tabular}{lccccccc}
\toprule
Post-trained & \textbf{Arc} & \textbf{Hellaswag} & \textbf{MMLU} & \textbf{TruthfulQA} & \textbf{WMT24++} & \textbf{WMT24++} & \textbf{Borda C} $\downarrow$ \\
    &  &  &  &  & \footnotesize{en→xx} & \footnotesize{xx→en} &  \\
    & (25-shot) & (10-shot) & (5-shot) & (0-shot) & (0-shot) & (0-shot) &  \\ \midrule
\textit{\textbf{Non-European}}  &  &  &  &  &  &  \\ \midrule
Gemma-2-9B-IT & \textbf{60.16} & \textbf{59.86} & \textbf{57.56} & \textbf{63.86} & \textbf{80.19} & \textbf{82.74} & \textbf{1.0} \\
LLaMa-3.1-8B-IT & 47.59 & 54.54 & 56.10 & 54.93 & 79.13 & 80.99 & 2.0 \\
Granite-3-8B-IT & 30.22 & 33.73 & 39.45 & 50.70 & 46.45 & 72.23 & 4.8 \\
Qwen-2.5-7B-IT & 35.78 & 39.75 & 54.30 & 54.29 & 56.33 & 77.74 & 3.0 \\
OLMo-2-7B-IT & 27.91 & 31.96 & 34.24 & 45.72 & 44.61 & 69.22 & 6.0 \\
Aya-Expanse-8B & 30.42 & 35.16 & 42.39 & 46.32 & 50.42 & 75.84 & 4.2 \\

\midrule
\textit{\textbf{European}}  &  &  &  &  &  &  \\ \midrule
Mistral-7B-IT & 51.52 & 50.35 & 47.71 & \textbf{59.74} & 71.31 & 79.08 & 3.5 \\
Ministral-8B-IT & 45.24 & 47.53 & 46.63 & 51.42 & 68.08 & 79.13 & 5.2 \\
Occiglot-7B-eu5-IT & 36.34 & 40.63 & 35.91 & 48.66 & 53.34 & 69.76 & 6.8 \\
Salamandra-7B-IT & 50.73 & 58.58 & 45.29 & 52.33 & 80.88 & 83.11 & 3.0 \\
Pharia-1-LLM-7B-C & 26.41 & 29.31 & 28.47 & 50.66 & 34.04 & 51.72 & 7.8 \\
Teuken-7B-IT-R-v0.4 & 51.84 & 57.09 & 29.15 & 56.41 & 73.60 & 74.88 & 4.2 \\
Teuken-7B-IT-C-v0.4 & 50.62 & 55.79 & 34.09 & 52.21 & 73.78 & 79.73 & 4.3 \\
\textbf{EuroLLM-9B-IT} & \textbf{55.42} & \textbf{58.78} & \textbf{54.07} & 57.82 & \textbf{83.75} & \textbf{83.51} & \textbf{1.2} \\

\bottomrule
\end{tabular}%
}
\caption{ Comparison of the post-trained versions of open-weight LLMs on Hungarian benchmarks. }
\label{tab:posttrained_hu}
\end{table}

\newpage

\subsubsection{Italian (it)}

\begin{table}[H]
\resizebox{\textwidth}{!}{%
\begin{tabular}{lcccccccc}
\toprule
Pre-trained & \textbf{Arc} & \textbf{Hellaswag} & \textbf{MMLU} & \textbf{TruthfulQA} & \textbf{MMLU-pro} & \textbf{MUSR} & \textbf{Borda C}  $\downarrow$ \\
 & (25-shot) & (10-shot) & (5-shot) & (0-shot) & (5-shot) & (0-shot) &  \\ \midrule
\textit{\textbf{Non-European}} &  &  &  &  &  &  &  \\ \midrule

Gemma-2-9B & \textbf{71.42} & \textbf{70.69} & 68.87 & 48.96 & 29.17 & 9.39 & \textbf{1.8} \\

LLaMa-3.1-8B & 62.63 & 65.53 & 59.40 & 48.39 & 18.57 & 6.09 & 4.2 \\

Granite-3-8B & 64.63 & 68.65 & 57.31 & 51.15 & 19.76 & \textbf{10.02} & 2.3 \\

Qwen-2.5-7B & 62.52 & 62.57 & \textbf{69.24} & \textbf{56.60} & \textbf{31.39} & 8.07 & 2.5 \\

OLMo-2-7B & 45.07 & 48.50 & 46.92 & 44.68 & 13.67 & 7.54 & 5.7 \\

Aya-23-8B & 59.20 & 66.31 & 51.41 & 48.70 & 13.90 & 6.68 & 4.5 \\

\midrule
\textit{\textbf{European}} &  &  &  &  &  &  &  \\ \midrule

Mistral-7B & 59.72 & 61.94 & 55.75 & 47.02 & 16.40 & 9.23 & 2.7 \\

Occiglot-7B-eu5 & 63.12 & \textbf{70.18} & 51.65 & 45.11 & 12.68 & 6.25 & 2.8 \\

Salamandra-7B & 63.95 & 66.95 & 43.40 & 45.91 & 4.99 & 4.36 & 3.3 \\

\textbf{EuroLLM-9B} & \textbf{69.25} & 68.79 & \textbf{56.98} & \textbf{50.92} & \textbf{17.64} & \textbf{14.05} & \textbf{1.2} \\

\bottomrule
\end{tabular}%
}
\caption{ Comparison of the pre-trained versions of open-weight LLMs on Italian benchmarks. }
\label{tab:pretrained_it}
\end{table}


\begin{table}[H]
\resizebox{\textwidth}{!}{%
\begin{tabular}{lccccccccc}
\toprule
Post-trained & \textbf{Arc} & \textbf{Hellaswag} & \textbf{MMLU} & \textbf{TruthfulQA} & \textbf{MUSR} & \textbf{MMLU-pro} & \textbf{WMT24++} & \textbf{WMT24++} & \textbf{Borda C} $\downarrow$ \\
    &  &  &  &  &  &  & \footnotesize{en→xx} & \footnotesize{xx→en} &  \\
    & (25-shot) & (10-shot) & (5-shot) & (0-shot) & (0-shot) & (5-shot) & (0-shot) & (0-shot) &  \\ \midrule
\textit{\textbf{Non-European}}  &  &  &  &  &  &  &  &  \\ \midrule
Gemma-2-9B-IT & \textbf{68.72} & 68.23 & 68.25 & 59.48 & 28.44 & 9.82 & 81.18 & 84.37 & \textbf{2.2} \\
LLaMa-3.1-8B-IT & 58.37 & 63.26 & 62.54 & 58.57 & 23.58 & 4.89 & 81.56 & 83.04 & 3.8 \\
Granite-3-8B-IT & 62.91 & \textbf{70.70} & 60.16 & \textbf{62.50} & 18.86 & \textbf{10.52} & 81.02 & 84.15 & 2.5 \\
Qwen-2.5-7B-IT & 57.52 & 61.37 & \textbf{68.97} & 61.72 & \textbf{29.63} & 6.81 & 79.31 & 83.96 & 3.4 \\
OLMo-2-7B-IT & 47.33 & 52.13 & 47.12 & 53.26 & 11.36 & 4.43 & 75.29 & 82.46 & 6.0 \\
Aya-Expanse-8B & 59.66 & 68.28 & 57.59 & 57.38 & 19.26 & 8.99 & \textbf{83.46} & \textbf{84.58} & 2.6 \\

\midrule
\textit{\textbf{European}}  &  &  &  &  &  &  &  &  \\ \midrule
Mistral-7B-IT & 61.83 & 62.13 & 53.99 & \textbf{62.38} & 17.32 & 7.01 & 78.26 & 82.31 & 4.5 \\
Ministral-8B-IT & 59.71 & 65.99 & \textbf{58.44} & 54.47 & \textbf{18.10} & 6.15 & 81.36 & 83.64 & 3.1 \\
Occiglot-7B-eu5-IT & 56.38 & \textbf{67.71} & 48.63 & 49.07 & 12.98 & 6.51 & 72.09 & 75.04 & 5.6 \\
Salamandra-7B-IT & 59.12 & 65.78 & 48.10 & 49.11 & 7.36 & 7.97 & 81.94 & 84.49 & 4.5 \\
Pharia-1-LLM-7B-C & 59.67 & 67.03 & 44.97 & 47.40 & 11.12 & \textbf{17.53} & 79.39 & 82.37 & 4.8 \\
Teuken-7B-IT-R-v0.4 & 57.98 & 63.18 & 37.06 & 57.11 & 9.17 & 2.71 & 79.89 & 79.19 & 6.1 \\
Teuken-7B-IT-C-v0.4 & 57.92 & 63.32 & 42.65 & 52.54 & 10.03 & 5.39 & 79.70 & 82.55 & 5.9 \\
\textbf{EuroLLM-9B-IT} & \textbf{63.72} & 67.45 & 57.35 & 54.22 & 15.98 & 8.83 & \textbf{84.43} & \textbf{84.98} & \textbf{2.0} \\

\bottomrule
\end{tabular}%
}
\caption{ Comparison of the post-trained versions of open-weight LLMs on Italian benchmarks. }
\label{tab:posttrained_it}
\end{table}

\newpage
\subsubsection{Latvian (lv)}

\begin{table}[H]
\resizebox{\textwidth}{!}{%
\begin{tabular}{lccccc}
\toprule
Pre-trained & \textbf{Arc} & \textbf{Hellaswag} & \textbf{MMLU} & \textbf{TruthfulQA} & \textbf{Borda C}  $\downarrow$ \\
 & (25-shot) & (10-shot) & (5-shot) & (0-shot) &  \\ \midrule
\textit{\textbf{Non-European}} &  &  &  &  &  \\ \midrule

Gemma-2-9B & \textbf{60.40} & \textbf{60.88} & \textbf{62.29} & 54.38 & \textbf{1.2} \\

LLaMa-3.1-8B & 42.97 & 44.18 & 48.20 & 51.44 & 2.5 \\

Granite-3-8B & 29.47 & 32.51 & 35.67 & 49.50 & 4.0 \\

Qwen-2.5-7B & 35.18 & 38.14 & 53.44 & \textbf{54.58} & 2.2 \\

OLMo-2-7B & 27.31 & 30.77 & 32.91 & 48.79 & 5.8 \\

Aya-23-8B & 28.67 & 32.39 & 33.70 & 48.60 & 5.2 \\

\midrule
\textit{\textbf{European}} &  &  &  &  &  \\ \midrule

Mistral-7B & 29.40 & 32.94 & 37.10 & 50.16 & 3.0 \\

Occiglot-7B-eu5 & 28.70 & 32.20 & 32.65 & 50.69 & 3.5 \\

Salamandra-7B & 55.52 & 58.34 & 40.89 & 49.99 & 2.5 \\

\textbf{EuroLLM-9B} & \textbf{62.88} & \textbf{63.48} & \textbf{54.24} & \textbf{56.47} & \textbf{1.0} \\

\bottomrule
\end{tabular}%
}
\caption{ Comparison of the pre-trained versions of open-weight LLMs on Latvian benchmarks. }
\label{tab:pretrained_lv}
\end{table}


\begin{table}[H]
\resizebox{\textwidth}{!}{%
\begin{tabular}{lccccccc}
\toprule
Post-trained & \textbf{Arc} & \textbf{Hellaswag} & \textbf{MMLU} & \textbf{TruthfulQA} & \textbf{WMT24++} & \textbf{WMT24++} & \textbf{Borda C} $\downarrow$ \\
    &  &  &  &  & \footnotesize{en→xx} & \footnotesize{xx→en} &  \\
    & (25-shot) & (10-shot) & (5-shot) & (0-shot) & (0-shot) & (0-shot) &  \\ \midrule
\textit{\textbf{Non-European}}  &  &  &  &  &  &  \\ \midrule
Gemma-2-9B-IT & \textbf{58.76} & \textbf{56.38} & \textbf{60.67} & \textbf{58.81} & \textbf{76.84} & \textbf{81.35} & \textbf{1.0} \\
LLaMa-3.1-8B-IT & 41.52 & 43.88 & 49.63 & 52.03 & 65.03 & 76.17 & 2.7 \\
Granite-3-8B-IT & 29.19 & 32.36 & 36.52 & 50.40 & 41.67 & 66.96 & 5.0 \\
Qwen-2.5-7B-IT & 36.06 & 38.39 & 52.91 & 52.99 & 52.44 & 76.59 & 2.7 \\
OLMo-2-7B-IT & 28.87 & 30.63 & 32.94 & 49.54 & 40.01 & 62.05 & 6.0 \\
Aya-Expanse-8B & 29.78 & 32.43 & 38.77 & 53.92 & 42.57 & 69.19 & 3.7 \\

\midrule
\textit{\textbf{European}}  &  &  &  &  &  &  \\ \midrule
Mistral-7B-IT & 29.40 & 32.84 & 34.27 & 51.66 & 42.41 & 66.68 & 6.3 \\
Ministral-8B-IT & 40.13 & 42.27 & 43.52 & 51.89 & 60.25 & 76.26 & 4.7 \\
Occiglot-7B-eu5-IT & 28.31 & 31.90 & 29.83 & 52.64 & 44.11 & 60.09 & 6.5 \\
Salamandra-7B-IT & 48.97 & 56.83 & 43.83 & 54.43 & 78.90 & 82.20 & 2.3 \\
Pharia-1-LLM-7B-C & 26.72 & 29.52 & 28.67 & 47.73 & 33.38 & 52.12 & 8.0 \\
Teuken-7B-IT-R-v0.4 & 47.69 & 53.81 & 35.94 & \textbf{57.18} & 74.60 & 73.02 & 3.5 \\
Teuken-7B-IT-C-v0.4 & 45.77 & 54.07 & 40.54 & 54.65 & 72.97 & 79.59 & 3.5 \\
\textbf{EuroLLM-9B-IT} & \textbf{57.57} & \textbf{60.78} & \textbf{52.79} & 55.64 & \textbf{84.40} & \textbf{83.02} & \textbf{1.2} \\

\bottomrule
\end{tabular}%
}
\caption{ Comparison of the post-trained versions of open-weight LLMs on Latvian benchmarks. }
\label{tab:posttrained_lv}
\end{table}

\newpage
\subsubsection{Lithuanian (lt)}

\begin{table}[H]
\resizebox{\textwidth}{!}{%
\begin{tabular}{lccccc}
\toprule
Pre-trained & \textbf{Arc} & \textbf{Hellaswag} & \textbf{MMLU} & \textbf{TruthfulQA} & \textbf{Borda C}  $\downarrow$ \\
 & (25-shot) & (10-shot) & (5-shot) & (0-shot) &  \\ \midrule
\textit{\textbf{Non-European}} &  &  &  &  &  \\ \midrule

Gemma-2-9B & \textbf{62.86} & \textbf{62.47} & \textbf{63.15} & \textbf{53.48} & \textbf{1.0} \\

LLaMa-3.1-8B & 44.32 & 44.99 & 48.23 & 48.61 & 2.8 \\

Granite-3-8B & 30.06 & 32.91 & 35.88 & 47.67 & 5.0 \\

Qwen-2.5-7B & 35.33 & 37.47 & 52.04 & 52.08 & 2.8 \\

OLMo-2-7B & 26.74 & 30.75 & 32.59 & 46.43 & 6.0 \\

Aya-23-8B & 32.23 & 37.64 & 39.00 & 49.85 & 3.5 \\

\midrule
\textit{\textbf{European}} &  &  &  &  &  \\ \midrule

Mistral-7B & 28.91 & 34.10 & 37.48 & 46.38 & 3.2 \\

Occiglot-7B-eu5 & 27.66 & 32.62 & 32.64 & 46.61 & 3.8 \\

Salamandra-7B & 56.69 & 59.16 & 41.45 & 48.63 & 2.0 \\

\textbf{EuroLLM-9B} & \textbf{64.29} & \textbf{63.66} & \textbf{54.47} & \textbf{53.37} & \textbf{1.0} \\

\bottomrule
\end{tabular}%
}
\caption{ Comparison of the pre-trained versions of open-weight LLMs on Lithuanian benchmarks. }
\label{tab:pretrained_lt}
\end{table}


\begin{table}[H]
\resizebox{\textwidth}{!}{%
\begin{tabular}{lccccccc}
\toprule
Post-trained & \textbf{Arc} & \textbf{Hellaswag} & \textbf{MMLU} & \textbf{TruthfulQA} & \textbf{WMT24++} & \textbf{WMT24++} & \textbf{Borda C} $\downarrow$ \\
    &  &  &  &  & \footnotesize{en→xx} & \footnotesize{xx→en} &  \\
    & (25-shot) & (10-shot) & (5-shot) & (0-shot) & (0-shot) & (0-shot) &  \\ \midrule
\textit{\textbf{Non-European}}  &  &  &  &  &  &  \\ \midrule
Gemma-2-9B-IT & \textbf{58.67} & \textbf{58.75} & \textbf{61.90} & \textbf{59.11} & \textbf{79.11} & \textbf{80.52} & \textbf{1.0} \\
LLaMa-3.1-8B-IT & 42.06 & 44.09 & 49.53 & 52.72 & 67.34 & 75.12 & 2.3 \\
Granite-3-8B-IT & 29.81 & 33.06 & 36.07 & 48.78 & 43.03 & 66.22 & 5.0 \\
Qwen-2.5-7B-IT & 35.14 & 37.83 & 52.41 & 53.84 & 53.16 & 73.57 & 3.0 \\
OLMo-2-7B-IT & 28.84 & 30.88 & 34.02 & 45.70 & 42.89 & 62.53 & 6.0 \\
Aya-Expanse-8B & 33.24 & 36.99 & 42.88 & 52.05 & 54.90 & 74.27 & 3.7 \\

\midrule
\textit{\textbf{European}}  &  &  &  &  &  &  \\ \midrule
Mistral-7B-IT & 29.15 & 33.43 & 35.44 & 49.71 & 45.24 & 67.26 & 6.0 \\
Ministral-8B-IT & 41.39 & 43.33 & 42.32 & 51.43 & 60.62 & 74.37 & 4.5 \\
Occiglot-7B-eu5-IT & 27.69 & 32.57 & 29.74 & 48.49 & 44.77 & 61.51 & 7.0 \\
Salamandra-7B-IT & 50.07 & 58.00 & 44.54 & 55.94 & 80.45 & 81.12 & 2.0 \\
Pharia-1-LLM-7B-C & 25.39 & 29.66 & 28.82 & 45.61 & 34.43 & 51.19 & 8.0 \\
Teuken-7B-IT-R-v0.4 & 49.31 & 54.69 & 36.46 & \textbf{57.01} & 71.26 & 68.98 & 3.5 \\
Teuken-7B-IT-C-v0.4 & 48.62 & 55.16 & 40.41 & 52.64 & 68.82 & 78.27 & 3.7 \\
EuroLLM-9B-IT & \textbf{57.95} & \textbf{61.37} & \textbf{53.86} & 54.23 & \textbf{84.43} & \textbf{81.81} & \textbf{1.3} \\

\bottomrule
\end{tabular}%
}
\caption{ Comparison of the post-trained versions of open-weight LLMs on Lithuanian benchmarks. }
\label{tab:posttrained_lt}
\end{table}

\newpage
\subsubsection{Polish (pl)}

\begin{table}[H]
\resizebox{\textwidth}{!}{%
\begin{tabular}{lccccc}
\toprule
Pre-trained & \textbf{Arc} & \textbf{Hellaswag} & \textbf{MMLU} & \textbf{TruthfulQA} & \textbf{Borda C}  $\downarrow$ \\
 & (25-shot) & (10-shot) & (5-shot) & (0-shot) &  \\ \midrule
\textit{\textbf{Non-European}} &  &  &  &  &  \\ \midrule

Gemma-2-9B & \textbf{69.13} & \textbf{67.22} & \textbf{66.32} & 53.38 & \textbf{1.2} \\

LLaMa-3.1-8B & 57.07 & 58.48 & 55.28 & 51.69 & 3.0 \\

Granite-3-8B & 41.93 & 44.10 & 45.03 & 49.87 & 5.0 \\

Qwen-2.5-7B & 53.37 & 54.31 & 64.14 & \textbf{53.96} & 2.8 \\

OLMo-2-7B & 34.82 & 38.41 & 39.57 & 47.51 & 6.0 \\

Aya-23-8B & 56.11 & 61.96 & 49.48 & 51.91 & 3.0 \\

\midrule
\textit{\textbf{European}} &  &  &  &  &  \\ \midrule

Mistral-7B & 56.44 & 55.39 & 51.73 & 50.54 & 2.5 \\

Occiglot-7B-eu5 & 48.03 & 48.92 & 43.05 & 43.03 & 4.0 \\

Salamandra-7B & 62.24 & 64.34 & 43.60 & 48.89 & 2.5 \\

\textbf{EuroLLM-9B} & \textbf{66.71} & \textbf{66.27} & \textbf{55.86} & \textbf{53.20} & \textbf{1.0} \\

\bottomrule
\end{tabular}%
}
\caption{ Comparison of the pre-trained versions of open-weight LLMs on Polish benchmarks. }
\label{tab:pretrained_pl}
\end{table}


\begin{table}[H]
\resizebox{\textwidth}{!}{%
\begin{tabular}{lccccccc}
\toprule
Post-trained & \textbf{Arc} & \textbf{Hellaswag} & \textbf{MMLU} & \textbf{TruthfulQA} & \textbf{WMT24++} & \textbf{WMT24++} & \textbf{Borda C} $\downarrow$ \\
    &  &  &  &  & \footnotesize{en→xx} & \footnotesize{xx→en} &  \\
    & (25-shot) & (10-shot) & (5-shot) & (0-shot) & (0-shot) & (0-shot) &  \\ \midrule
\textit{\textbf{Non-European}}  &  &  &  &  &  &  \\ \midrule
Gemma-2-9B-IT & \textbf{64.32} & \textbf{64.18} & \textbf{65.11} & \textbf{62.42} & 80.42 & 81.55 & \textbf{1.3} \\
LLaMa-3.1-8B-IT & 50.58 & 56.19 & 57.67 & 58.13 & 77.69 & 79.41 & 3.2 \\
Granite-3-8B-IT & 41.60 & 45.90 & 46.67 & 53.31 & 59.04 & 77.56 & 5.2 \\
Qwen-2.5-7B-IT & 48.51 & 54.68 & 63.91 & 59.39 & 70.65 & 80.64 & 3.2 \\
OLMo-2-7B-IT & 35.27 & 39.62 & 40.36 & 51.75 & 59.94 & 75.89 & 5.8 \\
Aya-Expanse-8B & 54.63 & 63.88 & 54.69 & 57.43 & \textbf{83.01} & \textbf{82.28} & 2.3 \\

\midrule
\textit{\textbf{European}}  &  &  &  &  &  &  \\ \midrule
Mistral-7B-IT & 58.31 & 55.89 & 50.06 & \textbf{61.72} & 73.22 & 78.92 & 3.0 \\
Ministral-8B-IT & 50.89 & 53.40 & 47.75 & 55.00 & 70.88 & 79.31 & 4.7 \\
Occiglot-7B-eu5-IT & 42.86 & 46.65 & 40.22 & 47.47 & 58.80 & 69.97 & 6.7 \\
Salamandra-7B-IT & 55.52 & 63.40 & 47.24 & 52.89 & 80.37 & 81.74 & 3.3 \\
Pharia-1-LLM-7B-C & 27.29 & 31.56 & 29.57 & 44.02 & 37.21 & 60.59 & 8.0 \\
Teuken-7B-IT-R-v0.4 & 55.54 & 60.84 & 37.64 & 54.80 & 71.83 & 74.49 & 4.5 \\
Teuken-7B-IT-C-v0.4 & 53.24 & 59.14 & 38.01 & 54.19 & 71.13 & 79.65 & 4.7 \\
\textbf{EuroLLM-9B-IT} & \textbf{60.10} & \textbf{64.13} & \textbf{55.54} & 55.34 & \textbf{84.24} & \textbf{81.93} & \textbf{1.2} \\

\bottomrule
\end{tabular}%
}
\caption{ Comparison of the post-trained versions of open-weight LLMs on Polish benchmarks. }
\label{tab:posttrained_pl}
\end{table}

\newpage
\subsubsection{Portuguese (pt)}

\begin{table}[H]
\resizebox{\textwidth}{!}{%
\begin{tabular}{lcccccccc}
\toprule
Pre-trained & \textbf{Arc} & \textbf{Hellaswag} & \textbf{MMLU} & \textbf{TruthfulQA} & \textbf{MMLU-pro} & \textbf{MUSR} & \textbf{Borda C}  $\downarrow$ \\
 & (25-shot) & (10-shot) & (5-shot) & (0-shot) & (5-shot) & (0-shot) &  \\ \midrule
\textit{\textbf{Non-European}} &  &  &  &  &  &  &  \\ \midrule

Gemma-2-9B & \textbf{71.55} & \textbf{71.14} & 68.09 & 47.78 & 29.31 & \textbf{11.58} & \textbf{1.8} \\

LLaMa-3.1-8B & 61.94 & 66.16 & 59.61 & 46.67 & 19.98 & 8.00 & 4.0 \\

Granite-3-8B & 63.66 & 69.14 & 57.77 & 48.27 & 20.44 & 9.16 & 2.7 \\

Qwen-2.5-7B & 64.29 & 65.50 & \textbf{69.73} & \textbf{56.52} & \textbf{31.57} & 9.09 & 2.2 \\

OLMo-2-7B & 48.82 & 50.38 & 48.58 & 44.26 & 13.23 & 2.45 & 6.0 \\

Aya-23-8B & 60.74 & 67.54 & 51.67 & 47.84 & 14.08 & 4.14 & 4.3 \\

\midrule
\textit{\textbf{European}} &  &  &  &  &  &  &  \\ \midrule

Mistral-7B & 61.35 & 62.38 & 55.97 & 44.91 & \textbf{17.54} & 11.77 & 2.2 \\

Occiglot-7B-eu5 & 57.82 & 61.47 & 48.97 & 42.05 & 11.54 & 4.96 & 3.5 \\

Salamandra-7B & 64.62 & 67.63 & 43.88 & 42.97 & 5.43 & 2.94 & 3.2 \\

\textbf{EuroLLM-9B} & \textbf{70.01} & \textbf{68.67} & \textbf{56.90} & \textbf{49.43} & 17.28 & \textbf{15.08} & \textbf{1.2} \\

\bottomrule
\end{tabular}%
}
\caption{ Comparison of the pre-trained versions of open-weight LLMs on Portuguese benchmarks. }
\label{tab:pretrained_pt}
\end{table}


\begin{table}[H]
\resizebox{\textwidth}{!}{%
\begin{tabular}{lccccccccc}
\toprule
Post-trained & \textbf{Arc} & \textbf{Hellaswag} & \textbf{MMLU} & \textbf{TruthfulQA} & \textbf{MUSR} & \textbf{MMLU-pro} & \textbf{WMT24++} & \textbf{WMT24++} & \textbf{Borda C} $\downarrow$ \\
    &  &  &  &  &  &  & \footnotesize{en→xx} & \footnotesize{xx→en} &  \\
    & (25-shot) & (10-shot) & (5-shot) & (0-shot) & (0-shot) & (5-shot) & (0-shot) & (0-shot) &  \\ \midrule
\textit{\textbf{Non-European}}  &  &  &  &  &  &  &  &  \\ \midrule
Gemma-2-9B-IT & \textbf{69.88} & 67.89 & 63.63 & \textbf{63.61} & 18.52 & 8.87 & 79.11 & 83.56 & \textbf{2.4} \\
LLaMa-3.1-8B-IT & 57.34 & 62.40 & 62.03 & 57.40 & 23.73 & 6.22 & 79.27 & 82.40 & 4.0 \\
Granite-3-8B-IT & 62.28 & \textbf{70.64} & 60.63 & 60.08 & 20.01 & \textbf{12.50} & 79.86 & 83.35 & \textbf{2.4} \\
Qwen-2.5-7B-IT & 59.42 & 63.61 & \textbf{68.88} & 62.05 & \textbf{28.26} & 6.78 & 78.62 & 83.32 & 3.1 \\
OLMo-2-7B-IT & 49.86 & 52.96 & 48.59 & 49.50 & 11.69 & 5.09 & 75.52 & 82.18 & 6.0 \\
Aya-Expanse-8B & 59.44 & 68.68 & 58.10 & 56.20 & 18.99 & 7.87 & \textbf{81.44} & \textbf{84.13} & 2.8 \\

\midrule
\textit{\textbf{European}}  &  &  &  &  &  &  &  &  \\ \midrule
Mistral-7B-IT & 63.01 & 62.22 & 53.78 & \textbf{61.38} & 17.62 & 7.05 & 76.61 & 81.71 & 4.2 \\
Ministral-8B-IT & 60.28 & 65.69 & 57.53 & 55.18 & \textbf{20.08} & 9.56 & 79.94 & 83.18 & 2.4 \\
Occiglot-7B-eu5-IT & 54.00 & 59.54 & 45.09 & 48.46 & 10.83 & 4.20 & 68.93 & 73.78 & 6.9 \\
Salamandra-7B-IT & 58.27 & 66.96 & 47.38 & 51.12 & 7.10 & 9.33 & 80.34 & 83.99 & 4.0 \\
Pharia-1-LLM-7B-C & 58.02 & 64.62 & 44.37 & 47.81 & 10.37 & \textbf{15.11} & 76.27 & 81.25 & 5.4 \\
Teuken-7B-IT-R-v0.4 & 58.64 & 63.95 & 38.68 & 49.28 & 8.87 & 2.78 & 77.02 & 77.38 & 6.2 \\
Teuken-7B-IT-C-v0.4 & 57.53 & 63.34 & 41.24 & 52.14 & 9.20 & 5.39 & 78.52 & 82.04 & 5.5 \\
\textbf{EuroLLM-9B-IT} & \textbf{64.23} & \textbf{68.37} & \textbf{56.63} & 54.65 & 15.23 & 12.60 & \textbf{82.03} & \textbf{84.17} & \textbf{1.8} \\

\bottomrule
\end{tabular}%
}
\caption{ Comparison of the post-trained versions of open-weight LLMs on Portuguese benchmarks. }
\label{tab:posttrained_pt}
\end{table}

\newpage
\subsubsection{Romanian (ro)}

\begin{table}[H]
\resizebox{\textwidth}{!}{%
\begin{tabular}{lccccc}
\toprule
Pre-trained & \textbf{Arc} & \textbf{Hellaswag} & \textbf{MMLU} & \textbf{TruthfulQA} & \textbf{Borda C}  $\downarrow$ \\
 & (25-shot) & (10-shot) & (5-shot) & (0-shot) &  \\ \midrule
\textit{\textbf{Non-European}} &  &  &  &  &  \\ \midrule

Gemma-2-9B & \textbf{69.45} & \textbf{67.81} & \textbf{67.77} & 51.81 & \textbf{1.8} \\

LLaMa-3.1-8B & 57.97 & 59.78 & 57.94 & 51.82 & 2.8 \\

Granite-3-8B & 45.11 & 45.59 & 46.90 & 52.48 & 4.2 \\

Qwen-2.5-7B & 51.27 & 49.04 & 64.81 & \textbf{57.05} & 2.8 \\

OLMo-2-7B & 40.04 & 42.69 & 44.09 & 48.99 & 6.0 \\

Aya-23-8B & 57.87 & 63.83 & 50.48 & 51.60 & 3.5 \\

\midrule
\textit{\textbf{European}} &  &  &  &  &  \\ \midrule

Mistral-7B & 55.79 & 54.87 & 53.68 & \textbf{53.89} & 2.2 \\

Occiglot-7B-eu5 & 49.32 & 48.81 & 43.42 & 47.58 & 3.8 \\

Salamandra-7B & 62.24 & 64.55 & 43.23 & 48.87 & 2.8 \\

\textbf{EuroLLM-9B} & \textbf{65.60} & \textbf{66.50} & \textbf{56.89} & 52.82 & \textbf{1.2} \\

\bottomrule
\end{tabular}%
}
\caption{ Comparison of the pre-trained versions of open-weight LLMs on Romanian benchmarks. }
\label{tab:pretrained_ro}
\end{table}


\begin{table}[H]
\resizebox{\textwidth}{!}{%
\begin{tabular}{lccccccc}
\toprule
Post-trained & \textbf{Arc} & \textbf{Hellaswag} & \textbf{MMLU} & \textbf{TruthfulQA} & \textbf{WMT24++} & \textbf{WMT24++} & \textbf{Borda C} $\downarrow$ \\
    &  &  &  &  & \footnotesize{en→xx} & \footnotesize{xx→en} &  \\
    & (25-shot) & (10-shot) & (5-shot) & (0-shot) & (0-shot) & (0-shot) &  \\ \midrule
\textit{\textbf{Non-European}}  &  &  &  &  &  &  \\ \midrule
Gemma-2-9B-IT & \textbf{65.42} & 63.59 & \textbf{66.72} & \textbf{61.29} & 81.76 & 83.74 & \textbf{1.5} \\
LLaMa-3.1-8B-IT & 53.76 & 57.59 & 59.39 & 57.91 & 80.29 & 82.18 & 3.0 \\
Granite-3-8B-IT & 46.02 & 46.58 & 50.33 & 57.79 & 61.77 & 79.80 & 5.2 \\
Qwen-2.5-7B-IT & 48.93 & 49.28 & 64.33 & 58.90 & 68.53 & 81.65 & 3.5 \\
OLMo-2-7B-IT & 41.09 & 44.16 & 45.47 & 50.55 & 70.18 & 80.31 & 5.5 \\
Aya-Expanse-8B & 57.32 & \textbf{64.68} & 56.48 & 55.96 & \textbf{84.85} & \textbf{84.67} & 2.3 \\

\midrule
\textit{\textbf{European}}  &  &  &  &  &  &  \\ \midrule
Mistral-7B-IT & 57.82 & 54.45 & 51.66 & \textbf{62.54} & 73.64 & 80.60 & 3.5 \\
Ministral-8B-IT & 51.98 & 52.57 & 53.46 & 54.02 & 71.53 & 81.03 & 4.7 \\
Occiglot-7B-eu5-IT & 44.64 & 47.54 & 40.61 & 52.06 & 62.82 & 71.63 & 6.5 \\
Salamandra-7B-IT & 56.30 & 62.97 & 48.37 & 49.92 & 82.43 & 84.25 & 3.3 \\
Pharia-1-LLM-7B-C & 31.18 & 33.12 & 33.90 & 48.16 & 38.69 & 67.62 & 8.0 \\
Teuken-7B-IT-R-v0.4 & 55.66 & 60.22 & 36.44 & 57.03 & 79.51 & 76.29 & 4.2 \\
Teuken-7B-IT-C-v0.4 & 53.99 & 59.57 & 40.39 & 53.06 & 79.09 & 82.17 & 4.5 \\
EuroLLM-9B-IT & \textbf{61.30} & \textbf{64.68} & \textbf{57.27} & 54.89 & \textbf{85.67} & \textbf{84.96} & \textbf{1.3} \\

\bottomrule
\end{tabular}%
}
\caption{ Comparison of the post-trained versions of open-weight LLMs on Romanian benchmarks. }
\label{tab:posttrained_ro}
\end{table}

\newpage
\subsubsection{Slovak (sk)}

\begin{table}[H]
\resizebox{\textwidth}{!}{%
\begin{tabular}{lccccc}
\toprule
Pre-trained & \textbf{Arc} & \textbf{Hellaswag} & \textbf{MMLU} & \textbf{TruthfulQA} & \textbf{Borda C}  $\downarrow$ \\
 & (25-shot) & (10-shot) & (5-shot) & (0-shot) &  \\ \midrule
\textit{\textbf{Non-European}} &  &  &  &  &  \\ \midrule

Gemma-2-9B & \textbf{66.69} & \textbf{65.21} & \textbf{65.61} & 50.14 & \textbf{1.2} \\

LLaMa-3.1-8B & 52.09 & 53.29 & 54.31 & 49.70 & 2.5 \\

Granite-3-8B & 45.11 & 48.51 & 47.25 & 47.25 & 4.0 \\

Qwen-2.5-7B & 45.88 & 47.12 & 62.57 & \textbf{50.98} & 2.8 \\

OLMo-2-7B & 29.65 & 34.13 & 37.53 & 42.64 & 6.0 \\

Aya-23-8B & 43.50 & 48.95 & 44.96 & 45.35 & 4.5 \\

\midrule
\textit{\textbf{European}} &  &  &  &  &  \\ \midrule

Mistral-7B & 45.24 & 46.42 & 48.32 & 43.26 & 2.8 \\

Occiglot-7B-eu5 & 40.23 & 43.10 & 41.29 & 42.40 & 4.0 \\

Salamandra-7B & 60.43 & 62.28 & 43.24 & 44.88 & 2.2 \\

\textbf{EuroLLM-9B} & \textbf{66.28} & \textbf{66.03} & \textbf{55.96} & \textbf{52.67} & \textbf{1.0} \\

\bottomrule
\end{tabular}%
}
\caption{ Comparison of the pre-trained versions of open-weight LLMs on Slovak benchmarks. }
\label{tab:pretrained_sk}
\end{table}


\begin{table}[H]
\resizebox{\textwidth}{!}{%
%
%
}
\caption{ Comparison of the post-trained versions of open-weight LLMs on Slovak benchmarks. }
\label{tab:posttrained_sk}
\end{table}

\newpage
\subsubsection{Slovenian (sl)}

\begin{table}[H]
\resizebox{\textwidth}{!}{%
%
%
}
\caption{ Comparison of the pre-trained versions of open-weight LLMs on Slovenian benchmarks. }
\label{tab:pretrained_sl}
\end{table}


\begin{table}[H]
\resizebox{\textwidth}{!}{%
%
%
}
\caption{ Comparison of the post-trained versions of open-weight LLMs on Slovenian benchmarks. }
\label{tab:posttrained_sl}
\end{table}

\newpage
\subsubsection{Spanish (es)}

\begin{table}[H]
\resizebox{\textwidth}{!}{%
%
%
}
\caption{ Comparison of the pre-trained versions of open-weight LLMs on Spanish benchmarks. }
\label{tab:pretrained_es}
\end{table}


\begin{table}[H]
\resizebox{\textwidth}{!}{%
%
%
}
\caption{ Comparison of the post-trained versions of open-weight LLMs on Spanish benchmarks. }
\label{tab:posttrained_es}
\end{table}

\newpage
\subsubsection{Swedish (sv)}

\begin{table}[H]
\resizebox{\textwidth}{!}{%
%
%
}
\caption{ Comparison of the pre-trained versions of open-weight LLMs on Swedish benchmarks. }
\label{tab:pretrained_sv}
\end{table}


\begin{table}[H]
\resizebox{\textwidth}{!}{%
%
%
}
\caption{ Comparison of the post-trained versions of open-weight LLMs on Swedish benchmarks. }
\label{tab:posttrained_sv}
\end{table}

\newpage
\subsection{Additional Languages}
\label{sec:others_lang_by_lang}

Results for pre-trained and post-trained models across eight languages which are not in the 24 official European languages (Catalan, Ukrainian, Arabic, Chinese, Hindi, Russian, Japanese, and Korean) are presented in the following tables. Among EU-made models, EuroLLM-9B demonstrates consistently superior performance across all these languages. In comparison with non-EU models, EuroLLM-9B has a comparable performance to Gemma-2-9B while surpassing the capabilities of other LLMs in the evaluation.

\subsubsection{Catalan (ca)}
\begin{table}[H]
\footnotesize
\setlength{\tabcolsep}{5.25ex}

\caption{Comparison of the pre-trained versions of open-weight LLMs on Catalan benchmarks.}
\end{table}


\begin{table}[H]
\resizebox{\textwidth}{!}{%
%
%
}
\caption{Comparison of the post-trained versions of open-weight LLMs on Catalan benchmarks.}
\end{table}

\newpage
\subsubsection{Ukrainian (uk)}
\begin{table}[H]
\footnotesize
\setlength{\tabcolsep}{5.25ex}
%
\caption{Comparison of the pre-trained versions of open-weight LLMs on Ukrainian benchmarks.}
\end{table}


\begin{table}[H]
\resizebox{\textwidth}{!}{%
%
}
\caption{Comparison of the post-trained versions of open-weight LLMs on Ukrainian benchmarks.}
\end{table}

\newpage

\subsubsection{Arabic (ar)}
\begin{table}[H]
\footnotesize
\setlength{\tabcolsep}{5.25ex}
%
\caption{Comparison of the pre-trained versions of open-weight LLMs on Arabic benchmarks.}
\end{table}

\begin{table}[H]
\resizebox{\textwidth}{!}{%
%
%
}
\caption{Comparison of the post-trained versions of open-weight LLMs on Arabic benchmarks.}
\end{table}

\newpage

\subsubsection{Chinese (zh)}
\begin{table}[H]
\footnotesize
\setlength{\tabcolsep}{3.5ex}
%
\caption{Comparison of the pre-trained versions of open-weight LLMs on Arabic benchmarks.}
\end{table}

\begin{table}[H]
\resizebox{\textwidth}{!}{%
%
%
}
\caption{Comparison of the post-trained versions of open-weight LLMs on Arabic benchmarks.}
\end{table}

\newpage

\subsubsection{Hindi (hi)}
\begin{table}[H]
\footnotesize
\setlength{\tabcolsep}{2.15ex}
%
\caption{Comparison of the pre-trained versions of open-weight LLMs on Hindi benchmarks.}
\end{table}

\begin{table}[H]
\resizebox{\textwidth}{!}{%
%
%
}
\caption{Comparison of the post-trained versions of open-weight LLMs on Hindi benchmarks.}
\end{table}

\newpage

\subsubsection{Russian (ru)}
\begin{table}[H]
\footnotesize
\setlength{\tabcolsep}{2.15ex}
%
\caption{Comparison of the pre-trained versions of open-weight LLMs on Russian benchmarks.}
\end{table}

\begin{table}[H]
\resizebox{\textwidth}{!}{%
%
%
}
\caption{Comparison of the post-trained versions of open-weight LLMs on Russian benchmarks.}
\end{table}

\newpage

\subsubsection{Japanese (ja)}
\begin{table}[H]
\footnotesize
\setlength{\tabcolsep}{7.7ex}
%
\caption{Comparison of the pre-trained versions of open-weight LLMs on Japanese benchmarks.}
\end{table}

\begin{table}[H]
\resizebox{\textwidth}{!}{%
%
%
}
\caption{Comparison of the post-trained versions of open-weight LLMs on Japanese benchmarks.}
\end{table}

\newpage

\subsubsection{Korean (ko)}
\begin{table}[H]
\footnotesize
\setlength{\tabcolsep}{7.75ex}
%
\caption{Comparison of the pre-trained versions of open-weight LLMs on Korean benchmarks.}
\end{table}

\begin{table}[H]
\resizebox{\textwidth}{!}{%
%
%
}
\caption{Comparison of the post-trained versions of open-weight LLMs on Korean benchmarks.}
\end{table}

\section{Prompt used to generate synthetic instructions}
\begin{figure*}
    \centering
\begin{tcolorbox}[
  width=\textwidth,
  colback=gray!10,
  colframe=black,
  boxrule=0.8mm,
  arc=2mm,
  title=\bfseries Prompt used to generate synthetic instructions,
  fonttitle=\bfseries,
]
\fontsize{9pt}{10pt}\selectfont
\#\# Definition of an ``Instruction'':

In the context of training a large language model (LLM) for an AI assistant, an "instruction" can be characterized as a specific directive or command given to the model to perform a particular task or behave in a certain way. An instruction for training an AI assistant is a clear, concise statement or question that:

\vspace{1\baselineskip}
1. Directs the model to perform a specific action or task

2. Provides context for the desired behavior or output

3. May include constraints or specific requirements

4. Is typically phrased in natural language

5. Can be general or domain-specific
\vspace{1\baselineskip}

Instructions are used to fine-tune the model's behavior, helping it understand and respond appropriately to various user inputs and scenarios. They are crucial in shaping the AI assistant's capabilities and ensuring it aligns with the intended use case. Instructions typically fall into one of the categories below:

\vspace{1\baselineskip}
1. Problem Solving: Coding, Mathematical reasoning, Knowledge and reasoning

2. Creative Tasks: Creative writing, Brainstorming

3. Information Processing: Summarization, Extraction, Classification, Translation

4. Question Answering: Open-ended, Closed-ended, Multiple Choice

5. Text Transformation: Rewriting

6. Roleplay and Simulation: Inhabiting a character/persona

7. Advisory: Asking for advice

8. Domain-Specific Knowledge: Humanity, history, and social studies, Other (specific domains could be added here as needed)

9. General / Miscellaneous

\vspace{1\baselineskip}
\#\# Task Description:

Considering the definition of an Instruction, analyze the following \textbf{\{language\}} web document and perform a three-step analysis:

$<$document$>$ \textbf{\{text\}} $<$document$/>$
\vspace{1\baselineskip}

Firstly, try to give a short summary in \textbf{\{language\}} of the document. This should help you better understand the document.

\vspace{1\baselineskip}
Secondly, using the knowledge in the document, create an instruction suitable for training an AI system, where the document provides the topic and, if possible, the answer. Prioritize complex instructions that require knowledge and reasoning to solve. Aim to create an instruction that falls into the \textbf{\{category\}} category.
Note that the instruction should be easy to understand **without requiring access to the document**. It should be clear and detailed enough for the model to comprehend the task without ambiguity, even without seeing the document.

\vspace{1\baselineskip}
Thirdly, categorize the instruction into one of the categories defined above.

\vspace{1\baselineskip}
Provide your final response in the following format:

\vspace{1\baselineskip}
Summary: $<$summary of the document in \textbf{\{language\}}$>$

Instruction: $<$\textbf{\{language\}} instruction for which the answer is in the document above$>$

Category: $<$one of the categories above$>$

\vspace{1\baselineskip}
NOTE: ``Summary:'', ``Instruction:'' and ``Category:'' keywords must be kept in english without any change while the generated summary and instruction should be in \textbf{\{language\}}. 

\vspace{1\baselineskip}
DO NOT provide the actual response to the instruction and make sure the instruction is in \textbf{\{language\}} even if the document is not in the same language/variant!

\end{tcolorbox}
\vspace{-1em} 
    \caption{Prompt used to generate synthetic instructions from monolingual web data. }
    \label{fig:synthethic-data-prompt}
\end{figure*}

\section{Prompt used to generate synthetic answers}
\begin{figure*}
    \centering
\begin{tcolorbox}[
  width=\textwidth,
  colback=gray!10,
  colframe=black,
  boxrule=0.8mm,
  arc=2mm,
  title=\bfseries Prompt used to generate synthetic answers,
  fonttitle=\bfseries,
]
\fontsize{9pt}{10pt}\selectfont
Consider the following \textbf{\{language\}} document and the \textbf{{language}} user’s instruction below:

$<$document$>$ \textbf{\{document\}} $<$/document$>$

\vspace{1\baselineskip}
$<$instruction$>$ \textbf{\{instruction\}} $</$instruction$>$

\vspace{1\baselineskip}
Given the above user’s instruction, and the context provided in the document, provide a response to the instruction in \textbf{{language}}. Your response should include, as much as possible, the information presented in the document while making the flow more clear, useful, relevant and providing a direct response to the instruction. It should be written to be impeccably tailored to the user’s instruction as if written by an AI Assistant, without extraneous information, reflecting expert knowledge, and demonstrating a high-quality, engaging, and insightful answer. Try not to add new facts that are not already in the document.

\vspace{1\baselineskip}
Since the user does not have access to the document, your answer cannot directly reference the document. The document serves only as supporting context for the instruction.

\vspace{1\baselineskip}
NOTE: Since the instruction is in \textbf{{language}}, your response should also be in \textbf{{language}}. Provide ONLY the response to the instruction.

\end{tcolorbox}
\vspace{-1em} 
    \caption{Prompt used to generate answers for the synthethic instructions created. The prompt uses the web document used to generate the instruction as context to better contextualize the provided answer.}
    \label{fig:synthethic-data-answer}
\end{figure*}

\end{document}